\documentclass[10pt,twocolumn,letterpaper]{article}

\usepackage[pagenumbers]{iccv} %

\definecolor{iccvblue}{rgb}{0.21,0.49,0.74}
\usepackage[pagebackref,breaklinks,colorlinks,allcolors=iccvblue]{hyperref}

\usepackage{booktabs}
\usepackage{multirow}
\usepackage{makecell}
\usepackage{colortbl}
\usepackage{pifont}%
\usepackage[accsupp]{axessibility}  

\newcommand{\cmark}{\color{green}{\ding{51}}}%
\newcommand{\xmark}{\color{red}{\ding{55}}}%

\definecolor{Gray}{gray}{0.93}
\definecolor{uclagold}{rgb}{1.0, 0.7, 0.0}
\definecolor{airforceblue}{rgb}{0.36, 0.54, 0.66}
\definecolor{rosegold}{rgb}{0.72, 0.43, 0.47}
\definecolor{pastelbrown}{rgb}{0.51, 0.41, 0.33}
\definecolor{isabelline}{rgb}{0.96, 0.94, 0.93}
\definecolor{macaroniandcheese}{rgb}{0.98, 0.89, 0.83}
\definecolor{wildblueyonder}{rgb}{0.85, 0.89, 0.95}
\definecolor{mediumtaupe}{rgb}{0.4, 0.3, 0.28}
\definecolor{bluegray}{rgb}{0.4, 0.6, 0.8}
\definecolor{celestialblue}{rgb}{0.29, 0.59, 0.82}
\definecolor{darkorange}{rgb}{1.0, 0.55, 0.0}
\definecolor{cadmiumred}{rgb}{0.89, 0.0, 0.13}
\definecolor{magnolia}{rgb}{0.97, 0.96, 1.0}
\definecolor{pastelblue}{rgb}{0.68, 0.78, 0.81}
\definecolor{persiangreen}{rgb}{0.0, 0.65, 0.58}
\definecolor{steelblue}{rgb}{0.27, 0.51, 0.71}
\definecolor{bluebell}{rgb}{0.64, 0.64, 0.82}
\definecolor{dimgray}{rgb}{0.41, 0.41, 0.41}
\definecolor{splashedwhite}{rgb}{1.0, 0.99, 1.0}
\definecolor{lavendergray}{rgb}{0.77, 0.76, 0.82}
\definecolor{lightgray}{rgb}{0.83, 0.83, 0.83}
\definecolor{lavendermist}{rgb}{0.9, 0.9, 0.98}
\definecolor{lightgreen}{HTML}{f8fcf4}
\definecolor{lightblue}{HTML}{dfebf7}

\newcommand{\Ours}{MM-Spatial\xspace}

\newcommand\blfootnote[1]{%
  \begingroup
  \renewcommand\thefootnote{}\footnote{#1}%
  \addtocounter{footnote}{-1}%
  \endgroup
}

\usepackage{tikz}
\newcommand*\circled[1]{%
    \tikz[baseline=(char.base)]{%
        \node[shape=circle,draw,inner sep=0pt,minimum size=11.pt] (char) {\footnotesize #1};%
    }%
}%

\urldef{\myurl}\url{https://github.com/apple/ml-cubifyanything}

\def\paperID{14773} %
\def\confName{ICCV}
\def\confYear{2025}

\title{\Ours: Exploring 3D Spatial Understanding in Multimodal LLMs}

\author{Erik Daxberger
\and
Nina Wenzel$^\dagger$
\and
David Griffiths$^\dagger$
\and
Haiming Gang
\and
Justin Lazarow
\and
\hspace{-3mm}
Gefen Kohavi
\and
\hspace{-3mm}
Kai Kang
\and
\hspace{-3mm}
Marcin Eichner
\and
\hspace{-3mm}
Yinfei Yang
\and
\hspace{-3mm}
Afshin Dehghan
\and
\hspace{-3mm}
Peter Grasch
\and
\emph{Apple}
}

\begin{document}

\twocolumn[{
\maketitle\centering
\vspace{-3mm}
\captionsetup{type=figure}
\includegraphics[width=1.0\textwidth]{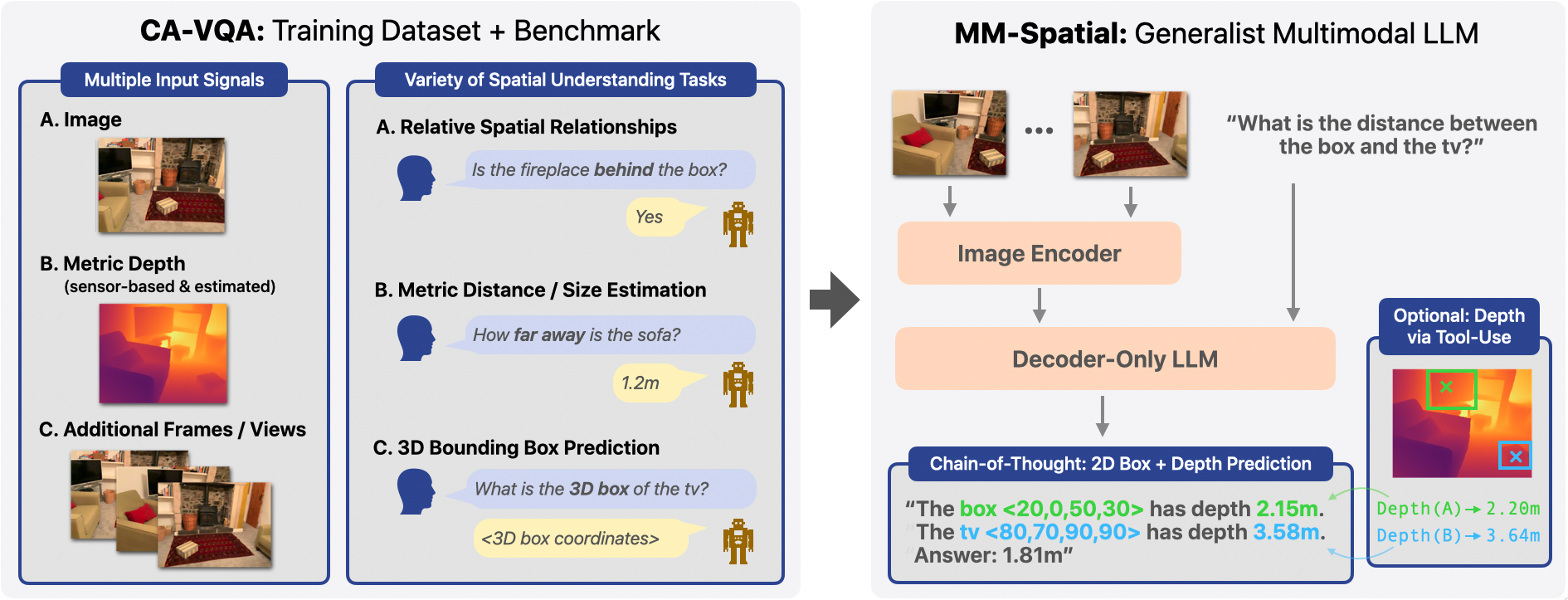}\vspace{-1mm}
\captionof{figure}{
(Left) We generate the \textbf{Cubify Anything VQA (CA-VQA)} dataset and benchmark, covering various 1) \emph{input signals}: single image, metric depth (sensor-based and estimated), multi-frame/-view, and 2) \emph{spatial understanding tasks}: e.g., relationship prediction, metric estimation, 3D grounding.
(Right) We train \textbf{\Ours}, a generalist multimodal LLM that excels at \emph{3D spatial understanding}. It supports \emph{Chain-of-Thought} spatial reasoning involving 2D grounding and depth estimation, and can also leverage \emph{depth input via tool-use}.
}
\label{fig:teaser}\vspace{5mm}
}]

\begin{abstract}
Multimodal large language models (MLLMs) excel at 2D visual understanding but remain limited in their ability to reason about 3D space.
In this work, we leverage large-scale high-quality 3D scene data with open-set annotations to introduce 1) a novel supervised fine-tuning dataset and 2) a new evaluation benchmark, focused on indoor scenes. Our Cubify Anything VQA (CA-VQA) data covers diverse spatial tasks including spatial relationship prediction, metric size and distance estimation, and 3D grounding. We show that CA-VQA enables us to train \Ours, a strong generalist MLLM that also achieves state-of-the-art performance on 3D spatial understanding benchmarks, including our own.
We show how incorporating metric depth and multi-view inputs (provided in CA-VQA) can further improve 3D understanding, and demonstrate that data alone allows our model to achieve depth perception capabilities comparable to dedicated monocular depth estimation models.
\small{\emph{\myurl}}

\phantom{\Ours: Exploring 3D Spatial Understanding in Multimodal LLMs}

\end{abstract}
    
\vspace*{-12mm}
\section{Introduction}
\label{sec:intro}
\blfootnote{$^\dagger$Equal contribution. E-Mail: \texttt{edaxberger@<company>.com}}%
Understanding object locations and spatial relationships in both 2D and 3D space is crucial for interpreting complex visual scenes.
While multimodal large language models (MLLMs) have achieved notable success for 2D visual tasks including referring and grounding \cite{you2023ferret,zhang2024ferret} and spatial relation prediction (e.g., \emph{``left''} vs.\ \emph{``right''}, \emph{``above''} vs.\ \emph{``below''}) \cite{cheng2024spatialrgpt}, they still struggle with 3D object perception tasks such as estimating 1) relative depth (\emph{``in front''} vs.\ \emph{``behind''}), 2) object distances or sizes in metric units (\emph{``A is 2.74m away / 1.32m wide.''}), and, ultimately, 3) precise 3D bounding boxes.
Yet, the ability to reason about objects in 3D scenes is not only a part of general visual comprehension, but is also foundational in domains like robotics and AR / VR, e.g., for navigation and manipulation tasks \cite{ma2024llmsstep3dworld}.

\begin{table*}[t!]
    \centering
    \resizebox{\linewidth}{!}{%
    \begin{tabular}{ll|cccccccccc}
         \toprule
         \multirow{3}{*}{\textbf{Dataset}} & \multirow{3}{*}{\textbf{Data source(s)}} & \multirow{3}{*}{\makecell{High-quality\\3D Ground-truth}} & \multicolumn{2}{c}{Depth maps} & \multirow{3}{*}{\makecell{Multi-view\\images}} & \multicolumn{3}{c}{Tasks} & \multicolumn{2}{c}{Splits} & \multirow{3}{*}{Public}\\
         \cmidrule(lr){4-5} \cmidrule(lr){7-9} \cmidrule(lr){10-11}
         & & & Sensor & Monoc. & & Relation & Metric & 3D Ground. & Train & Eval &\\
         \midrule
         
         \multicolumn{1}{c}{\textit{Training Datasets}}\\
         \midrule
         OpenSpatialDataset \cite{cheng2024spatialrgpt} & OpenImages \cite{kuznetsova2020open} & \xmark & \xmark & \cmark & \xmark & \cmark & \cmark & \xmark & \cmark & \xmark & \cmark\\
         SpatialQA-E \cite{cai2024spatialbot} & Robot manipulation images \cite{cai2024spatialbot} & \xmark & \xmark & \xmark & \xmark & \cmark & \xmark & \xmark & \cmark & \xmark & \cmark\\
         OpenSpaces \cite{remyxai2024openspaces} & The Cauldron \cite{laurenccon2025matters} & \xmark & \xmark & \xmark & \xmark & \cmark & \cmark & \xmark & \cmark & \cmark & \cmark\\
         Spatial Aptitude Training \cite{ray2024sat} & ProcTHOR-10K \cite{deitke2022procthor} & synthetic & \xmark & \xmark & \xmark & \cmark & \xmark & \xmark & \cmark & \cmark & \cmark\\
         RoboSpatial \cite{song2024robospatial} & Multiple 3D datasets \cite{dai2017scannet,chang2017matterport3d,wald2019rio,tyree20226dof,fang2020graspnet,wang2024embodiedscan}  & \cmark & \xmark & \xmark & \xmark & \cmark & \xmark & \xmark & \cmark & \cmark & \cmark\\
         EmbSpatial \cite{du2024embspatial} & Multiple 3D datasets \cite{dai2017scannet,chang2017matterport3d,kolve2017ai2}  & \cmark & \xmark & \xmark & \xmark & \cmark & \xmark & \xmark & \cmark & \cmark & \cmark\\
         \rowcolor{lightgray} 
         SpatialQA \cite{cai2024spatialbot} & Multiple image datasets & subset & \xmark & \cmark & \xmark & \cmark & \cmark & \xmark & \cmark & \xmark & \xmark\\
         \rowcolor{lightgray} 
         Spatial-VQA \cite{chen2024spatialvlm} & Web-crawled images & \xmark & \xmark & \xmark & \xmark & \cmark & \cmark & \xmark & \cmark & \xmark & \xmark\\
         \rowcolor{lightgray} 
         LV3D \cite{cho2024cubellm} & Multiple datasets & subset & \xmark & \xmark & \xmark & \xmark & \xmark & \cmark & \cmark & \xmark & \xmark\\
         
         \midrule  
         \multicolumn{1}{c}{\textit{Evaluation Benchmarks}}\\
         \midrule
         SpatialRGPT-Bench \cite{cheng2024spatialrgpt} & Omni3D \cite{brazil2023omni3d} & \cmark & \xmark & \cmark & \xmark & \cmark & \cmark & \xmark & \xmark & \cmark & \cmark\\
         CV-Bench \cite{tong2024cambrian} & ADE20k \cite{zhou2017ade20k}, COCO \cite{cocodataset}, Omni3D \cite{brazil2023omni3d} & subset & \xmark & \xmark & \xmark & \cmark & \xmark & \xmark & \xmark & \cmark & \cmark\\
         3DSRBench \cite{ma20243dsrbench} & COCO \cite{cocodataset}, HSSD \cite{khanna2023hssd} & synthetic & \xmark & \xmark & \cmark & \cmark & \xmark & \xmark & \xmark & \cmark & \cmark\\
         VSI-Bench \cite{yang2024thinking} & ScanNet/++ \cite{dai2017scannet,yeshwanth2023scannet++}, ARKitScenes \cite{dehghan2021arkitscenes} & \cmark & \xmark & \xmark & \cmark & \cmark & \cmark & \xmark & \xmark & \cmark & \cmark\\
         Q-Spatial \cite{liaos2024reasoning} & ScanNet \cite{dai2017scannet} & \cmark & \xmark & \xmark & \xmark & \xmark & \cmark & \xmark & \xmark & \cmark & \cmark\\
         ScanRefer \cite{chen2020scanrefer} & ScanNet \cite{dai2017scannet} & \cmark & \cmark & \xmark & \cmark & \xmark & \xmark & \cmark & \xmark & \cmark & \cmark\\
         Nr3D / Sr3D \cite{achlioptas2020referit3d} & ScanNet \cite{dai2017scannet} & \cmark & \cmark & \xmark & \cmark & \xmark & \xmark & \cmark & \xmark & \cmark & \cmark\\
         SpatialBench \cite{cai2024spatialbot} & subset is from MME \cite{fu2023mme} & \xmark & \xmark & \cmark & \xmark & \cmark & \xmark & \xmark & \xmark & \cmark & \cmark\\
         Rel3D \cite{goyal2020rel3d} & ShapeNet \cite{chang2015shapenet}, YCB \cite{calli2015ycb} & \cmark & \xmark & \xmark & \xmark & \cmark & \xmark & \xmark & \xmark & \cmark & \cmark\\
         
         \midrule  
         \rowcolor{lightgreen} 
         \textbf{CA-VQA (ours)} & CA-1M \cite{lazarow2024cubify} / ARKitScenes \cite{dehghan2021arkitscenes} & \cmark & \cmark & \cmark & \cmark & \cmark & \cmark & \cmark & \cmark & \cmark & \cmark\\
         
         \bottomrule
    \end{tabular}
    }
    \caption{\textbf{3D Spatial Dataset Overview.} Comparison of object-centric 3D spatial MLLM datasets to CA-VQA (in gray: non-public ones). CA-VQA is the first dataset that is based on high-quality 3D ground truth, includes depth maps (both from sensors and monocular) and multi-view images, covers a variety of tasks (relationships, metric estimation, 3D grounding), and has both an SFT dataset and benchmark.
    }
    \label{tab:dataset_comparison}
\end{table*}

There have been comparatively few works on 3D object perception with MLLMs \cite{chen2024spatialvlm,cheng2024spatialrgpt,cai2024spatialbot,cho2024cubellm,song2024robospatial,du2024embspatial}; moreover, they only consider a subset of tasks, and do not comprehensively assess depth and multi-view inputs.
To address these limitations and facilitate a more holistic exploration of 3D understanding in MLLMs, we make these contributions:

\begin{enumerate}
    \item We propose a new data generation pipeline that leverages high-quality 3D scene data to produce image-text QA pairs for 3D object perception. We apply this pipeline to CA-1M \cite{lazarow2024cubify} to generate \emph{Cubify Anything VQA (CA-VQA)}, a new spatial understanding dataset for MLLM fine-tuning, covering diverse indoor scenes. As additional inputs, CA-VQA uniquely includes multi-view images and
    different types of metric depth maps, both sensor-based and SOTA monocular (estimated) depth.
    \item We release a new spatial understanding benchmark derived from CA-VQA. Compared to existing benchmarks, ours 1) includes diverse tasks (incl.\ relative and metric distance / size estimation and 3D grounding), 2) provides rich input signals (multi-view and depth), and 3) is less susceptible to language priors (i.e., more vision-reliant) and hence more challenging.
    We show that even SO\-TA models such as GPT-4o struggle on our benchmark.
    \item We run extensive experiments illustrating the benefits of CA-VQA as a testbed for spatial perception research.
    We show that 1) we can train \Ours, a generalist MLLM achieving SOTA on spatial understanding benchmarks (CV-Bench, SpatialRGPT-Bench, CA-VQA), while retaining performance on other tasks (incl.\ general, knowledge, text-rich); 2) using multi-view and depth inputs further enhances 3D understanding; 3) MLLMs can achieve strong monocular depth estimation via SFT.
    We also study the efficacy of different depth maps, the impact of full encoding vs.\ tool-use for leveraging depth, and indoor-to-outdoor scene generalization.
\end{enumerate}

\begin{figure*}[t!]
\centering
{
\includegraphics[width=1.0\textwidth]{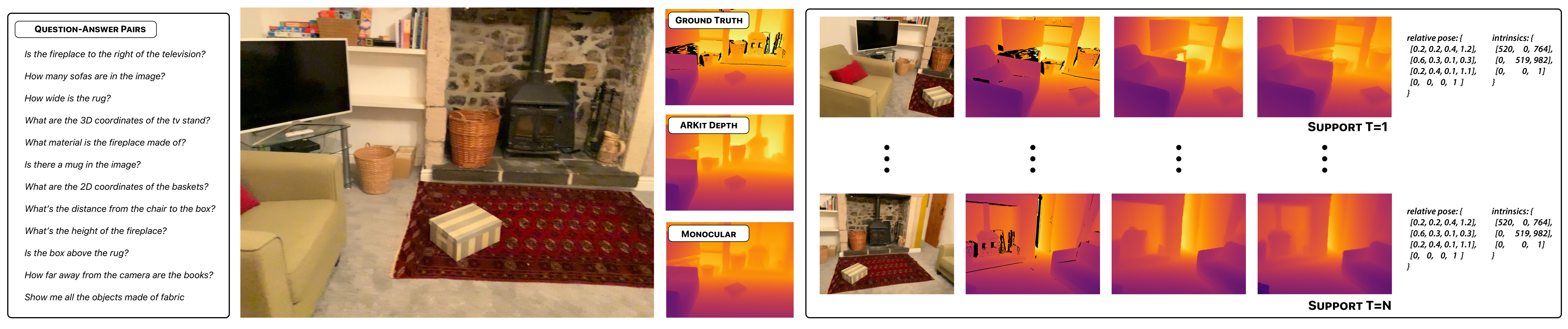}
}
\caption{\textbf{CA-VQA Data Example.} Example of a single sample from our dataset. Each reference frame has between 0-4 multi-view support frames. All frames (reference and support) come with three metric depth maps: Ground truth (FARO laser), ARKit Depth (LiDAR-fused) and Monocular (DepthPro). Each support frame contains the relative pose from the reference image, along with camera intrinsics.}
\label{fig:dataset_sample}
\end{figure*}

\section{Related Work}
\label{sec:related_work}

\subsection{General MLLMs}

MLLMs \cite{openai2024gpt4vision,islam2024gpt,team2023gemini,li2024multimodal,huang2023language} have attracted significant research focus, tracing back to Frozen~\cite{tsimpoukelli2021multimodal} and Flamingo~\cite{alayrac2022flamingo,awadalla2023openflamingo}, with more recent works such as LLaVA~\cite{llava} and MiniGPT-4~\cite{zhu2023minigpt} introducing visual instruction tuning. The rise of open-source MLLMs has led to models rivaling SOTA commercial offerings like GPT-4o on certain tasks. Notable examples include Emu2~\cite{sun2023generative,sun2024generative}, VILA~\cite{lin2024vila}, Idefics2/3~\cite{laurenccon2025matters,laurenccon2024building}, and Qwen2-VL~\cite{bai2023qwen}, among others.

MLLMs research has explored several fronts: ($i$) scaling up the pre-training data~\cite{lin2024vila,mckinzie2024mm1,xue2024xgen,awadalla2024mint,li2024omnicorpus} and supervised fine-tuning data~\cite{hu2024mplug,tang2024textsquare,laurenccon2025matters,tong2024cambrian}; ($ii$) enhancing high-resolution image comprehension~\cite{lin2023sphinx,li2023monkey,liu2024llavanext,dong2024internlm,gao2024sphinx,ge2024convllava,chen2024dragonfly,zhang2024beyond,xu2024llava,li2024mini}; ($iii$) studying various vision encoders~\cite{tong2024eyes,shi2024eagle,chen2024cloc,fini2024aimv2} and vision-language connectors~\cite{cha2024honeybee,yao2024dense,li2024tokenpacker,cai2024matryoshka}; ($iv$) using mixture-of-experts~\cite{lin2024moe,li2024cumo}; ($v$) extending models to region-level~\cite{wang2023visionllm,zhao2023bubogpt,zang2023contextual,peng2023kosmos,chen2023shikra,zhang2023gpt4roi,you2023ferret,zhang2024ferret} and pixel-level~\cite{lai2024lisa,rasheed2024glamm,yuan2024osprey,ren2024pixellm} understanding, multi-image reasoning~\cite{jiang2024mantis,li2024llava}, UI comprehension~\cite{you2024ferret,li2024ferretui2,hong2024cogagent}, and video understanding~\cite{lin2023video,xu2024pllava,xu2024slowfast,ye2024mmego}, among others.

\subsection{3D Spatial Understanding with MLLMs}

To complement work on (primarily) 2D spatial relationships / reasoning \cite{zhang2024countercurate,yang2019spatialsense,shiri2024empirical,zhang2025do,agrawal2023stupd,pantazopoulos2024lost,ramakrishnan2024does,Liu2022VisualSR,su202225VRD,kamath2023whatsup,mayer2025ivispar}, recent research has aimed to also enable 3D reasoning with MLLMs, roughly split into two directions.
Firstly, works focusing on scene-level 3D understanding (i.e., scene captioning and VQA) by enabling MLLMs to process representations of entire scenes, often leveraging multiple views and depth information.
This includes ScanReason \cite{zhu2024scanreason}, 3D-CLR \cite{hong20233d}, 3D-LLM \cite{hong20233dllm}, ConceptGraphs \cite{gu2024conceptgraphs}, LLaVA-3D \cite{zhu2024llava3d}, Scene-LLM \cite{fu2024scenellm}, M3DBench \cite{li2023m3dbench}, Video-3D LLM \cite{zheng2024video}, LSceneLLM \cite{zhi2024lscenellm}, and 3DGraphLLM \cite{zemskova20243dgraphllm}.

Secondly, works focusing on object-level 3D spatial perception.
Spatial-VLM \cite{chen2024spatialvlm} and Cube-LLM \cite{cho2024cubellm} both use vanilla image-based VLMs (without any explicit 3D input) to address spatial relationship and metric distance estimation (Spatial-VLM)  as well as 3D grounding (Cube-LLM).
SpatialRGPT \cite{cheng2024spatialrgpt} and VCoder \cite{jain2024vcoder} encode (relative) depth maps as additional inputs via the image encoder plus a dedicated depth connector.
SpatialBot \cite{cai2024spatialbot} instead leverages depth maps via tool-use \cite{schick2023toolformer} by training the model to query the depth value at a given coordinate.
In this work we build on these ideas, and further compare the utility of depth maps collected with dedicated specialized hardware to those derived from monocular depth estimators. We also study the benefits of providing additional views (images) to the model, i.e., frames preceding the main image in the video.

We provide an overview of previous SFT datasets and benchmarks for object-centric 3D spatial understanding in \cref{tab:dataset_comparison}, highlighting the novelty and unique characteristics of our proposed CA-VQA dataset, to be detailed next.

\section{Data}
\label{sec:data}

We build upon the Cubify Anything 1M (CA-1M) \cite{lazarow2024cubify} dataset, which contains exhaustive 3D bounding boxes (gravity-aligned 7-DOF boxes with yaw orientation) for every object in the ARKitScenes \cite{dehghan2021arkitscenes} dataset. Additionally, we provide human-labeled annotations for each object consisting of an open-set label ($\sim$3.3k unique noun labels for $\sim$350k objects), material, primary color, and shape.

\subsection{Data Generation Pipeline}
\label{subsec:data_generation_pipeline}
We generate QA pairs from these annotations as follows:
\begin{itemize}[topsep=0pt, leftmargin=*]
    \item \textbf{Frame Sub-sampling.}
    To reduce the data volume and redundancy, we sub-sample the videos at 1 FPS for the training set and at 0.1 FPS for the evaluation benchmark.
    \item \textbf{3D Ground Truth Processing.} For each frame, we transform the 3D boxes from world to camera space using pose $Rt_{i}$. In contrast to CA-1M, 1) we include all boxes visible from the current view, irrespective of distance; 2) we do not clip boxes to the visible part, but rather store amodal 3D coordinates. We also construct a point cloud based on the ground truth depth map and camera intrinsics.
    \item \textbf{QA Pair Generation.} Based on our 3D and semantic annotations, we automatically generate template-based QA pairs (both open-ended and multi-choice ones), without any human supervision. We consider a variety of spatial task categories, detailed in \cref{subsec:spatial_task_categories} and App.\ \ref{app:data_details}. We further ensure that questions are unambiguous. For example, asking \textit{``What is the distance to the chair?''}, is only a valid question is there is a single instance of a chair.
    \item \textbf{Blind Filtering.} \cite{chen2024mmstar} found that many samples of multimodal benchmarks can be solved without vision input due to the strong language priors of MLLMs. To reduce such bias we follow \cite{chen2024mmstar} and remove all benchmark examples which are correctly answered \emph{blindly} by at least three out of seven judges: GPT-4 \cite{achiam2023gpt4}, GPT-4V \cite{openai2024gpt4vision}, GPT-4o \cite{islam2024gpt}, Phi-3-Vision-4B \cite{abdin2024phi}, LLaVA-OneVision-7B \cite{li2024llava_onevision}, SpatialRGPT-VILA1.5-8B \cite{cheng2024spatialrgpt}, and our \Ours-3B. App.\ \ref{app:blind_filtering_analysis} demonstrates the effectiveness of this strategy.
\end{itemize}

\noindent Overall, we obtain $\sim$10M QA pairs over 220K frames from 2K videos for the CA-VQA training set, and $\sim$62K QA pairs over 2.6K frames from 265 videos for the evaluation benchmark.
\cref{fig:qa_examples} shows QA examples from CA-VQA.

\subsection{Spatial Task Categories}
\label{subsec:spatial_task_categories}

CA-VQA covers the spatial task categories outlined below. App.\ \ref{app:data_details} provides further details on the QA definitions.
\newline \noindent \textbf{Counting} (\textit{``How many X are there?''}). Answers are computed by counting the 3D boxes of the given object class.
\newline \noindent \textbf{Viewpoint-dependent} (\textit{``Is X behind Y?''}). Answers are based on the 2D / 3D boxes and depend on the camera pose.
\newline \noindent \textbf{Metric regression} (\textit{``How far away is X from Y / the camera?'' ``How wide / tall is X?''}) Answers to size questions are computed using the 3D bounding box. Answers to distance questions are computed based on the object point clouds; we reject samples for which the 3D boxes overlap.
\newline \noindent \textbf{2D referring/grounding}. We use 2D bounding boxes computed by projecting the 3D bounding boxes to image space. 
\newline \noindent \textbf{3D referring/grounding}. We use the 3D boxes in CA-1M.
\newline \noindent \textbf{Binary} (e.g., \textit{``Is X taller than Y?''}). This covers \emph{viewpoint-dependent} and (relative) \emph{regression} questions, as well as \emph{object presence} questions (\textit{``Is X present in the image?''}).
\newline \noindent \textbf{Multi-choice}. We also formulate multi-choice QAs covering the other categories (except for 2D / 3D grounding).

\noindent \textbf{External benchmark templates.}
We also generate examples using the QA templates proposed in CV-Bench \cite{tong2024cambrian} and SpatialRGPT-Bench \cite{cheng2024spatialrgpt}.
This removes potential instruction following issues due to differences in QA formulations, hence allowing us to faithfully evaluate our model's actual spatial understanding ability on those benchmarks.

\subsection{Multi-view and Metric Depth Data} 
\label{subsec:mv_and_depth}
\cref{fig:dataset_sample} visualizes the multi-view images and depth maps.

\noindent \textbf{Multi-view.} For each reference frame $I_t$, we sample $N \leq 4$ preceding support frames $I_{t-1, ..., t-N}$, which are
triggered when camera pose $Rt_i$ has angular movement of $\geq 15^{\circ}$ or movement of $\geq 30\text{cm}$ from the current key frame $Rt_b$.

\noindent \textbf{Metric Depth.} For each frame (ref.\ \& support), we provide:
\begin{itemize}[topsep=0pt, leftmargin=*]
    \item \textbf{Ground truth depth} acquired from a high-precision stationary FARO laser scanner, and rendered to each frame using the Barrabandi pipeline from ARKitScenes \cite{dehghan2021arkitscenes}.
    \item \textbf{ARKit Depth} provided by the ARKit framework. It utilizes the iPad Pro's on-device sparse LiDAR sensor and color image to produce a per-pixel dense depth map.
    \item \textbf{Monocular depth} generated using DepthPro \cite{bochkovskii2024depth}, a state-of-the-art monocular metric depth estimation model.
\end{itemize}

\begin{figure}[t!]
\centering
\includegraphics[width=0.47\textwidth]{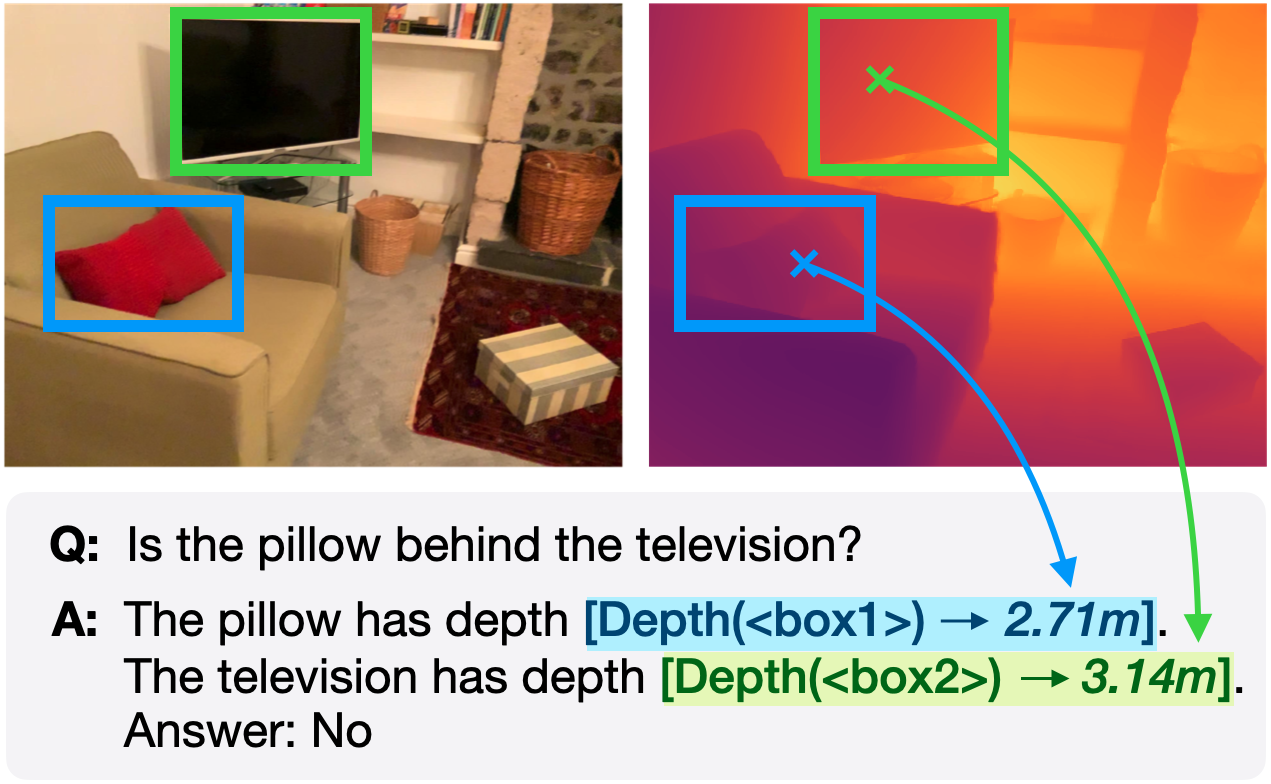}
\caption{Example of leveraging depth maps via tool-use. The model predicts the objects' 2D bounding boxes and function calls, receives the \emph{tool outputs} (which is the median depth value within the box, marked with an $\mathbf{\times}$), and finally reasons about the answer.}
\label{fig:depth_tool}
\end{figure}

\noindent \textbf{Depth: Chain-of-Thought (CoT) / Tool-Use.}
\cref{sec:experiments} will explore different ways to use the metric depth information.
As an alternative to encoding the full 2D depth maps with the model (see \cref{subsec:model_architecture}), we investigate a simpler approach that uses the individual depth values of the given objects as text. To this end, we prepare step-by-step examples involving textual GT depth in the format illustrated in \cref{fig:depth_tool}.
At test time, the depth values are then obtained either via 1) tool-use \cite{schick2023toolformer,cai2024spatialbot} (see \cref{fig:depth_tool}), or 2) model prediction (CoT).

\section{Model}
\label{sec:model}

\begin{figure*}[t!]
\centering
\includegraphics[width=\textwidth]{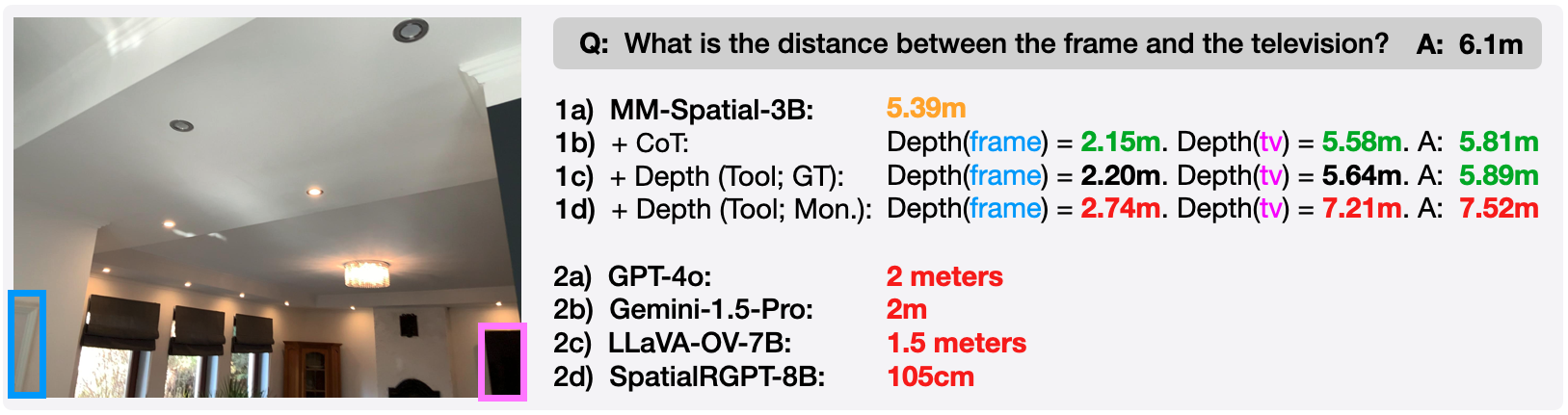}
\caption{\textbf{Qualitative Example.} We show the predictions of various models on a challenging example from our CA-VQA benchmark. Strong commercial (2a\&b) and research models (2c\&d) fail. \Ours (1a) is much better, and even more so with CoT enabled (1b), demonstrating our model's strong object grounding (see predicted 2D boxes in the image), depth estimation, and spatial reasoning ability. Accuracy improves further when leveraging ground-truth depth via tool-use (1c), although our CoT model's (1b) predictions are very close to that, for both the intermediate depth values and final answer; monocular estimated depth (1d) is less accurate and yields a worse result.}
\label{fig:qualitative_example}
\end{figure*}

\subsection{Model Architecture}
\label{subsec:model_architecture}

We use the MM1.5 architecture \cite{zhang2024mm1,mckinzie2024mm1} (focusing on the mobile-friendly 3B variant), comprising of a DFN-CLIP \cite{fang2023data,radford2021learning} image encoder and a decoder-only LLM backbone, which are bridged via a C-Abstractor \cite{cha2024honeybee}.
We use an image resolution of 672×672 during fine-tuning, and further increase the effective resolution by using (static) image splitting \cite{lin2023sphinx} with 4 sub-image splits (plus an overview image). 

\noindent We also consider variants of our model that incorporate either multiple views or depth maps as additional inputs:
\begin{itemize}[topsep=0pt, leftmargin=*]
    \item \textbf{Multi-view.}
    Our model supports multi-image input, allowing us to concatenate multiple views into sequences $I_{t-N}, ..., I_{t-1}, I_t$.
    In this multi-view setting, we only apply image splitting to the reference (final) image $I_t$.
    \item \textbf{Depth: Full Encoding.}
    We use the image encoder to encode the normalized and colorized depth maps (i.e., replicating the depth map along the channel dimension), and introduce a separate depth connector, following SpatialRGPT \cite{cheng2024spatialrgpt}. Notably, this approach is limited to using \emph{relative} (normalized) depth.
    We also explore using textual \emph{metric} depth in a purely data-driven way via Chain-of-Thought (CoT) or tool-use, see \cref{subsec:mv_and_depth,subsec:model_variants}.
\end{itemize}

\subsection{Data and Training}
\label{subsec:data_training}

We follow the 1) pre-training and 2) continual pre-training stages of MM1.5 \cite{zhang2024mm1}. For the 3) supervised fine-tuning (SFT) stage, we start from the MM1.5 single-image SFT mixture, which includes datasets across multiple categories: \emph{General} VQA, \emph{Knowledge} (math, code, science), \emph{Text-rich}, and 2D \emph{Referring \& Grounding} (VQA enriched with bounding boxes).
We then add our CV-VQA data within a new \emph{Spatial} category and select the mixture ratio based on the ablations discussed in App.\ \ref{app:ablation_mixture_weights}.
We use the same training hyperparameters as MM1.5~\cite{zhang2024mm1}, with \emph{unfrozen} image encoder and LLM.
We use AXLearn \cite{apple2024axlearn} for model training.

\section{Experiments}
\label{sec:experiments}

\begin{table}[t!]
    \centering
    \resizebox{\linewidth}{!}{%
    \begin{tabular}{l|ccccc|c}
         \toprule
         \multirow{3}{*}{\textbf{Model}} & \multicolumn{6}{c}{Benchmark Category Averages}\\
         \cmidrule{2-7}
         & Spatial & General & Knowl. & Text-rich & Ref./Ground & \textbf{Avg.}\\
         \midrule
         
         \rowcolor{lightblue} 
         MM1.5-3B \cite{zhang2024mm1} & 39.9 & 64.7 & 46.2 & 62.1 & 77.7 & 58.1\\
         \rowcolor{lightgreen} 
         \Ours-3B & 70.1 & 65.0 & 46.2 & 62.1 & 79.1 & 64.5\\
         
         \bottomrule
    \end{tabular}
    }
    \caption{\textbf{Benchmark Category Results}
    \Ours is a generalist MLLM that improves strongly on the \emph{Spatial} category while rivaling the MM1.5 baseline across the other task categories.
    }
    \label{tab:overview_benchmark_categories}
\end{table}

\subsection{Model Variants}
\label{subsec:model_variants}

We explore the following model variants in our study, leveraging the various input signals provided within CA-VQA:
\begin{itemize}
    \item \textbf{\Ours.} Trained on single-view RGB inputs, without depth information. This is our baseline model.
    \item \textbf{\Ours + Multi-view.} Trained on multi-view RGB inputs as described in \cref{subsec:model_architecture}. We use up to four support frames plus one reference frame.
    We additionally provide the camera intrinsics and pose information for each view (relative to the reference view) as JSON strings to the model (see \cref{fig:dataset_sample}), interleaved with the images.
    \item \textbf{\Ours + Depth (Tool).}
    Trained on single-view RGB plus textual metric depth, with tool-use at test time, as described in \cref{subsec:mv_and_depth,fig:depth_tool}.
    This approach relies solely on data, using the same model as \textbf{\Ours}.
    We denote the depth source as \textbf{(Tool; GT)} for ground truth / FARO depth or \textbf{(Tool; Mon.)} for monocular depth.
    \item \textbf{\Ours + CoT.}
    Trained like \textbf{\Ours + Depth (Tool)}, but \emph{without} using the depth tool at test time. Instead, the model predicts the metric depth values on its own based solely on the image input, producing a Chain-of-Thought (CoT) \cite{wei2022chain} style response as in \cref{fig:teaser,fig:depth_tool}.
    \item \textbf{\Ours + Depth (Encoded).} Trained on single-view RGB inputs plus fully encoded depth maps, as described in \cref{subsec:model_architecture}.
    The depth source is denoted as above.    
    \item \textbf{\Ours (Blind eval).}
    Trained like \textbf{\Ours}, but evaluated with text input only (i.e., without image input).
\end{itemize}

\begin{table*}[t!]
    \centering
    \resizebox{\linewidth}{!}{%
    \begin{tabular}{cl|rrrrrrrr|c}
         \toprule
         & \multirow{4}{*}{\textbf{Model}} & \multirow{3}{*}{Binary} & \multirow{3}{*}{Count.} & \multicolumn{2}{c}{Grounding} & \multirow{3}{*}{Multi-c.} & \multicolumn{3}{c|}{Regression (Metric Estimation)} & \multirow{4}{*}{\textbf{Average}}\\
         \cmidrule{5-6} \cmidrule{8-10}
         & & & & \multicolumn{1}{c}{2D} & \multicolumn{1}{c}{3D} & & Ego-Dist. & Obj.-Dist. & Obj.-Size & \\
         \cmidrule{3-10} 
         & & Acc & Acc & AP@50 & AP@15 & Acc & \multicolumn{3}{c|}{Acc @ 10\% Relative Error ($\ell_1$)} & \\
         
         \midrule

         \circled{1} & GPT-4 \cite{achiam2023gpt4} (\texttt{gpt-4-0613}) & 9.6 & 8.5 & 0.0 & 0.0 & 9.6 & 6.2 & 6.2 & 5.8 & 5.7\\
         \circled{2} & GPT-4V~\cite{openai2024gpt4vision} (\texttt{gpt-4-turbo-2024-04-09}) & 39.2 & 63.3 & 5.8 & 0.0 & 32.9 & 11.4 & 9.3 & 10.1 & 21.5\\
         \circled{3} & GPT-4o~\cite{islam2024gpt} (\texttt{gpt-4o-2024-08-06}) & 44.2 & 69.0 & 0.0 & 0.0 & 36.6 & 11.7 & 10.0 & 11.0 & 22.8\\
         \circled{4} & Phi-3-Vision-4B \cite{abdin2024phi} & 52.3 & 45.7 & 7.8 & 0.0 & 32.2 & 6.6 & 4.4 & 6.1 & 19.4\\
         \circled{5} & LLaVA-OneVision-7B \cite{li2024llava} & 52.0 & 62.1 & 16.1 & 0.0 & 42.5 & 9.3 & 8.1 & 6.4 & 24.6\\
         \circled{6} & SpatialRGPT-VILA1.5-8B \cite{cheng2024spatialrgpt} & 53.6 & 68.8 & 5.5 & 0.0 & 37.2 & 10.5 & 8.7 & 7.0 & 23.9\\
         \rowcolor{lightblue}
         \circled{7} & MM1.5-3B \cite{zhang2024mm1} & 59.1 & 9.1 & 32.6 & 0.0 & 38.6 & 0.6 & 2.2 & 3.4 & 18.2\\
         \rowcolor{lightgreen}
         \circled{8} & \Ours-3B & 68.8 & 75.8 & 53.2 & 20.7 & 74.2 & 40.0 & 18.7 & 24.4 & 47.0\\
         \rowcolor{lightgreen}
         \circled{9} & \Ours-3B + CoT & 69.6 & 75.9 & 54.5 & 21.9 & 74.7 & 46.0 & 23.2 & 26.7 & 49.1\\
         \rowcolor{lightgreen}
         \circled{10} & \Ours-3B + Depth (Tool; Mon.) & 69.6 & 75.9 & 54.5 & 21.9 & 74.7 & 40.9 & 23.8 & 26.6 & 48.5\\
         \rowcolor{lightgreen}
         \rowcolor{lightgreen}
         \circled{11} & \Ours-3B + Multi-view + CoT & 69.2 & 76.1 & 55.0 & 23.6 & 75.3 & 46.1 & 24.0 & 28.2 & 49.7\\
         \rowcolor{lightgreen}
         \rowcolor{lightgreen}
         \circled{12} & \Ours-3B + Multi-view + Depth (Tool; GT) & 69.2 & 76.1 & 55.0 & 23.6 & 75.3 & 65.8 & 27.2 & 27.3 & 52.4\\

         \midrule  
         \multicolumn{2}{c}{\textit{Specialist Models}}\\
         \midrule  

         \rowcolor{lightgreen}
         \circled{13} & \Ours-3B & 69.6 & 73.3 & 54.7 & 24.0 & 77.4 & 47.3 & 24.4 & 24.3 & 49.4\\
         \rowcolor{lightgreen}
         \circled{14} & \Ours-3B + CoT & 70.1 & 73.3 & 55.8 & 25.1 & 77.7 & 49.5 & 27.9 & 26.7 & 50.8\\
         \rowcolor{lightgreen}
         \circled{15} & \Ours-3B + Depth (Tool; Mon.) & 70.1 & 73.3 & 55.8 & 25.1 & 77.7 & 42.1 & 26.1 & 26.1 & 49.5\\
         \rowcolor{lightgreen}
         \circled{16} & \Ours-3B + Depth (Tool; GT) & 70.1 & 73.3 & 55.8 & 25.1 & 77.7 & 74.0 & 32.4 & 27.4 & 54.5\\
         \rowcolor{lightgreen}
         \circled{17} & \Ours-3B + Depth (Encoded; GT) & 69.5 & 73.1 & 55.6 & 24.2 & 78.3 & 48.3 & 25.4 & 24.5 & 49.9\\
         \rowcolor{lightgreen}
         \circled{18} & \Ours-3B + Depth (Encoded; GT) + CoT & 69.8 & 73.5 & 55.3 & 24.4 & 77.7 & 51.4 & 27.6 & 26.5 & 50.8\\
         \rowcolor{lightgreen}
         \circled{19} & \Ours-3B + Multi-view & 71.5 & 74.1 & 56.2 & 26.8 & 77.9 & 52.4 & 26.2 & 26.1 & 51.4\\
         \rowcolor{lightgreen}
         \circled{20} & \Ours-3B + Multi-view + CoT & 71.1 & 73.8 & 57.2 & 27.5 & 78.9 & 55.2 & 29.7 & 28.6 & 52.7\\
         \rowcolor{lightgreen}
         \circled{21} & \Ours-3B + Multi-view + Depth (Tool; Mon.) & 71.1 & 73.8 & 57.2 & 27.5 & 78.9 & 42.1 & 26.7 & 26.7 & 50.5\\
         \rowcolor{lightgreen}
         \circled{22} & \Ours-3B + Multi-view + Depth (Tool; GT) & 71.1 & 73.8 & 57.2 & 27.5 & 78.9 & 73.1 & 33.0 & 28.1 & 55.3\\
         \rowcolor{lightgreen}
         \circled{23} & \Ours-3B (Blind eval) & 34.3 & 60.8 & 0.0 & 0.0 & 60.7 & 10.1 & 8.4 & 17.9 & 24.0\\

         \bottomrule
    \end{tabular}
    }
    \caption{\textbf{CA-VQA Results.} \Ours-3B significantly outperforms (much larger) top open-source and commercial models across all tasks, demonstrating its strong spatial understanding ability. Model performance is further improved by incorporating multi-view and/or depth as additional input signals, as well as by leveraging CoT, which relies on our model's ability to accurately estimate metric depth.}
    \label{tab:results_ca-vqa}
\end{table*}

\noindent Some variants naturally lend themselves to combinations, e.g., \textbf{\Ours + Multi-view + CoT} is a model that sees multiple views and responds with CoT style answers.

\subsection{Overview of Benchmark Category Results}
\label{subsec:benchmark_category_overview}

We aim to train a generalist MLLM that excels across a variety of tasks -- instead of a specialist model that \emph{only} excels at spatial reasoning. 
To this end, we follow MM1.5 \cite{zhang2024mm1} and evaluate our models across 24 multimodal benchmarks using an internal fork of lm-eval-harness~\cite{Gao_2023b}, covering the categories outlined in \cref{subsec:data_training}.
To evaluate 2D and 3D \emph{Spatial} understanding we use CV-Bench \cite{tong2024cambrian}, SpatialRGPT-Bench \cite{cheng2024spatialrgpt}, and our proposed CA-VQA benchmark.

\cref{tab:overview_benchmark_categories} shows results on aggregated metrics across the various benchmark categories.
\Ours significantly improves on the \emph{Spatial} category while maintaining performance competitive with MM1.5 across the other categories, suggesting that spatial reasoning can be improved without meaningful compromise. 
See App.\ \ref{app:additional_results} for additional discussion.
We now present an in-depth analysis of the \emph{Spatial} results, comparing \Ours with SOTA baselines.
We use the full data mixture in \cref{subsec:data_training} by default; some ablations use \emph{Specialist Models} trained only on CA-VQA.

\subsection{Results on our CA-VQA Benchmark}
\label{subsec:ca-vqa_results}

We assess the model variants outlined in \cref{subsec:model_architecture}, and also study the metric depth estimation ability of our CoT model. 
CA-VQA results are shown in \cref{tab:results_ca-vqa}; a qualitative example is shown in \cref{fig:qualitative_example}. We make the following observations:

\begin{itemize}[topsep=0pt, leftmargin=*]
    \item \textbf{\Ours vs.\ Baselines.} \Ours-3B \circled{8} substantially outperforms various (much larger) top open-source and commercial models \circled{1}-\circled{6}, incl.\ the SOTA GPT-4o model \circled{3}, demonstrating 1) their limitations in terms of spatial understanding and 2) the effectiveness of SFT on CA-VQA.
    Despite its focus on spatial reasoning, SpatialRGPT-VILA-1.5-8B \circled{6} underperforms, likely for a few reasons: 1) their OpenSpatialDataset (OSD) used for SFT leverages \emph{axis-aligned} 3D boxes pseudo-annotated on OpenImages \cite{kuznetsova2020open}, resulting in significant discrepancies in spatial concept definitions (especially for metric quantities) compared to the high-quality (yaw-)\emph{oriented} 3D boxes used for CA-VQA; see App.\ \ref{app:aabb_vs_obb} for further discussion on this difference; 2) in OSD, objects are referred to via segmentation masks or 2D bounding boxes (\emph{``How tall is Region [0] \texttt{<mask/box>}?''}), while CA-VQA simply uses class names (\emph{``How tall is the chair?''}); 3) SpatialRGPT is limited to using \emph{relative} depth due to its full depth encoding approach. Finally, SpatialRGPT / OSD lacks support for 3D grounding.
    \item \textbf{Blind vs.\ Vision evaluation.} To validate our blind filtering strategy (see \cref{subsec:data_generation_pipeline}) we note that GPT-4 \circled{1} performs poorly, whereas its vision counterpart, GPT-4V \circled{2}, performs better.
    \Ours (Blind) \circled{23}, which is trained on similar data, also performs poorly on most tasks. While the performance on \emph{Counting} and \emph{Multi-choice} is still acceptable -- likely due to inherent remaining biases such as the naturally skewed distribution of object counts -- providing vision input \circled{13} still further improves performance by $\sim$15 points.
    App.\ \ref{app:blind_filtering_analysis} provides further detailed analysis of the effectiveness of our blind filtering strategy. Overall, our results suggest that our benchmark is less susceptible to a strong language prior compared to, e.g., SpatialRGPT-Bench (see \cref{subsec:spatialrgpt-bench_results}).
    \item \textbf{Multi-view vs.\ Single-view.} Multi-view is consistently better (e.g.\ \circled{19} vs.\ \circled{13}), suggesting that our model can successfully use additional views to improve 3D perception.
    
\begin{table*}[t!]
    \centering
    \resizebox{\linewidth}{!}{%
    \begin{tabular}{cl|cc|c|ccc|ccc|c|c}
         \toprule
         & \multirow{4}{*}{\textbf{Model}} & \multicolumn{3}{c|}{2D Tasks (CV-Bench$^\text{2D}$)} & \multicolumn{7}{c|}{3D Tasks (CV-Bench$^\text{3D}$)} & \multirow{4}{*}{\makecell{\textbf{Average}\\(2D+3D)}}\\
         \cmidrule{3-12}
         & & \multirow{3}{*}{\makecell{Object\\Count}} & \multirow{3}{*}{\makecell{Spatial\\Relation.}} & \multirow{3}{*}{\makecell{\textbf{Average}\\(2D)}} & \multicolumn{3}{c|}{Depth Order} & \multicolumn{3}{c|}{Relative Distance} & \multirow{3}{*}{\makecell{\textbf{Average}\\(3D)}} &\\
         \cmidrule{6-11}
         & & & & & Indoor & Outdoor & \textbf{Avg.} & Indoor & Outdoor & \textbf{Avg.} & & \\
         
         \midrule
         \circled{1} & GPT-4V~\cite{openai2024gpt4vision,tong2024cambrian} & -- & -- & 64.3 & -- & -- & -- & -- & -- & -- & 73.8 & 69.1\\
         \circled{2} & LLaVA-NeXT-8B \cite{liu2024llavanext,tong2024cambrian} & -- & -- & 62.2 & -- & -- & -- & -- & -- & -- & 65.3 & 63.8\\
         \circled{3} & Cambrian-1-8B \cite{tong2024cambrian} & -- & -- & 72.3 & -- & -- & -- & -- & -- & -- & 72.0 & 72.2\\
         \circled{4} & Phantom-7B \cite{lee2024phantom} & -- & -- & -- & -- & -- & -- & -- & -- & -- & -- & 74.9\\
         \circled{5} & LLaVA-1.5-13B + SAT Dyn.\ \cite{ray2024sat} & 62.9 & 85.8 & 74.4 & -- & -- & 76.6 & -- & -- & 71.6 & 74.1 & 74.3\\
         \circled{6} & LLaVA-NeXT-34B \cite{liu2024llavanext,tong2024cambrian} & -- & -- & 73.0 & -- & -- & -- & -- & -- & -- & 74.8 & 73.9\\
         \circled{7} & Mini-Gemini-HD-34B \cite{li2024mini,tong2024cambrian} & -- & -- & 71.5 & -- & -- & -- & -- & -- & -- & 79.2 & 75.4\\
         \circled{8} & Cambrian-1-34B \cite{tong2024cambrian} & -- & -- & 74.0 & -- & -- & -- & -- & -- & -- & 79.7 & 76.9\\
         \rowcolor{lightblue} 
         \circled{9} & MM1.5-3B \cite{zhang2024mm1} & 58.6 & 64.5 & 61.3 & 67.0 & 71.5 & 68.5 & 68.3 & 69.0 & 68.5 & 68.5 & 64.9\\
         \rowcolor{lightgreen} 
         \circled{10} & \Ours-3B & \textbf{88.7} & 94.0 & 91.1 & 96.8 & 87.0 & 93.5 & 95.8 & 75.5 & 89.0 & 91.3 & 91.2\\
         \rowcolor{lightgreen} 
         \circled{11} & \Ours-3B + CoT & 88.1 & \textbf{96.2} & \textbf{91.8} & 96.5 & 88.0 & 93.7 & \textbf{98.5} & 78.0 & 91.7 & 92.7 & 92.3\\
         \rowcolor{lightgreen} 
         \circled{12} & \Ours-3B + Depth (Tool; Mon.) & 88.1 & \textbf{96.2} & \textbf{91.8} & \textbf{99.0} & \textbf{92.0} & \textbf{96.7} & 97.8 & \textbf{80.0} & \textbf{91.8} & \textbf{94.3} & \textbf{93.1}\\
         
         \midrule  
         \rowcolor{lightgreen} 
         \circled{13} & \Ours-3B (Specialist; Blind eval) & \textbf{92.0} & 59.1 & 77.1 & 64.5 & 50.0 & 59.7 & 59.0 & 51.5 & 56.5 & 58.1 & 67.6\\
         
         \bottomrule
    \end{tabular}
    }
    \caption{\textbf{CV-Bench Results.} \Ours-3B substantially outperforms the (much larger) SOTA models, with CoT and depth input further improving performance. It almost fully solves the indoor 3D tasks, while also excelling at the out-of-domain outdoor 3D tasks.}
    \label{tab:results_cvbench}
\end{table*}

\begin{table}[t!]
    \centering
    \resizebox{\linewidth}{!}{%
    \begin{tabular}{l|cc|cc}
         \toprule
         \multirow{2}{*}{\textbf{Model}} & \multicolumn{2}{c|}{Generalist} & \multicolumn{2}{c}{Specialist}\\
         \cmidrule{2-5}
         & $\delta_1 \uparrow$ & \emph{AbsRel} $\downarrow$ & $\delta_1 \uparrow$ & \emph{AbsRel} $\downarrow$\\
         
         \midrule
         ARKit Depth & 96.7 & 4.2 & 96.1 & 4.6\\
         Monocular (DepthPro \cite{bochkovskii2024depth}) & 82.2 & 13.4 & 82.0 & 13.6\\
         \rowcolor{lightgreen} 
         \Ours-3B + CoT & 84.6 & 12.8 & 89.4 & 11.0\\
         
         \bottomrule
    \end{tabular}
    }
    \caption{\textbf{Metric Depth Estimation Results.} We evaluate the metric depth estimates of our CoT model produced as part of its responses on the CA-VQA benchmark.
    We compare against the tool-use estimates based on Monocular (DepthPro \cite{bochkovskii2024depth}) and ARKit Depth, i.e., the median depth value within the 2D box predicted by \Ours + CoT (generalist \& specialist). 
    We report the $\delta_1$ (accuracy at 25\% relative error) and \emph{AbsRel} (absolute relative error) metrics \cite{ladicky2014pulling} commonly used in the depth estimation literature \cite{bochkovskii2024depth}, computed against GT FARO depth. \Ours + CoT outperforms DepthPro. LiDAR-derived ARKit Depth is best.
    }
    \label{tab:results_depth_estimation}
\end{table}

    \item \textbf{Multi-view \circled{19} vs.\ Single-view + Depth (Tool; GT) \circled{16}.} While multi-view is competitive overall, on \emph{Regression} using GT depth is much better. This suggests that using multiple views can partially compensate if depth sensors are not available, but on tasks that most directly rely on accurate depth (i.e., \emph{Regression}), GT depth is unmatched.
    \item \textbf{CoT vs.\ Direct Prediction.} CoT prediction consistently improves over direct prediction (e.g.\ \circled{9} vs.\ \circled{8}), suggesting that the additional multi-step supervision signal (incl.\ 2D object grounding and depth prediction) and/or leveraging more test-time compute benefits model accuracy.
    \item \textbf{Depth (GT): Tool-use vs.\ Full Encoding.} Full depth encoding \circled{17} performs much worse than tool-use \circled{16}, and is only slightly better than the RGB-only baseline \circled{13}. Firsty, this highlights the effectiveness of the simple tool-use approach in utilizing \emph{metric} depth (while full encoding can only use \emph{relative} depth). Secondly, this indicates the difficulty of effectively encoding and interpreting full depth maps with an MLLM -- the simple architecture proposed by \cite{cheng2024spatialrgpt} may be too limited in this regard, suggesting that further research is required in this direction.
    \item \textbf{Depth (Tool): Ground Truth vs.\ Monocular.} 
    Despite DepthPro being a strong monocular depth estimator, its limitations are still apparent on our benchmark, especially for metric estimation. This is particularly visible on the \emph{Ego-Distance} task which most directly relies on accurate depth estimation: monocular depth \circled{15} 1) performs substantially worse than GT depth \circled{16}, and 2) even regresses compared to the RGB-only baseline \circled{13}, suggesting that our model itself can learn to accurately predict depth.
    \item \textbf{Depth-tool (Monocular) vs.\ CoT.} Our CoT approach \circled{14} (which requires \Ours to explicitly predict depth) performs better than tool-use with monocular depth \circled{15}, again most noticeable on \emph{Ego-Distance}.
    This again hints at MM-Spatial's strong inherent metric depth estimation ability, which we analyze in more detail below.
\end{itemize}

\begin{table*}[t!]
    \centering
    \resizebox{\linewidth}{!}{%
    \begin{tabular}{cll|cccccc|c|ccccc|c|c}
         \toprule
         & \multirow{3}{*}{\textbf{Model}} & \multirow{3}{*}{\textbf{Spatial SFT Data}} & \multicolumn{7}{c|}{Qualitative (Binary) Tasks} & \multicolumn{6}{c|}{Quantitative (Metric) Tasks} & \multirow{3}{*}{\textbf{Avg.}}\\
         \cmidrule{4-16}
         & & & \makecell{Below /\\Above} & \makecell{Left /\\Right} & \makecell{Big /\\Small} & \makecell{Tall /\\Short} & \makecell{Wide /\\ Thin} & \makecell{Behind\\/ Front} & \textbf{Avg.} & \makecell{Direct\\Dist.} & \makecell{Horizon.\\Dist.} & \makecell{Vertical\\Dist.} & Width & Height & \textbf{Avg.} & \\
         
         \midrule
         \circled{1} & GPT-4 \cite{achiam2023gpt4} & -- & 64.1 & 42.8 & 42.8 & 61.6 & 61.6 & 49.0 & 53.7 & 21.6 & 11.5 & 33.0 & 52.3 & 48.1 & 33.3 & 43.5 \\
         \circled{2} & GPT-4V~\cite{openai2024gpt4vision} & -- & 63.3 & 46.6 & 64.1 & 60.7 & 68.2 & 45.4 & 58.0 & 29.7 & 25.4 & 33.0 & 51.1 & 68.4 & 41.5 & 49.8\\
         \circled{3} & SpatialRGPT-7B (RGB-only) \cite{cheng2024spatialrgpt} & OSD \cite{cheng2024spatialrgpt} & \textbf{99.2} & 99.0 & 79.2 & 89.2 & 83.6 & 87.2 & 89.6 & 35.1 & 59.0 & 53.8 & 51.9 & 54.9 & 50.9 & 70.3\\
         \circled{4} & SpatialRGPT-7B \cite{cheng2024spatialrgpt} & OSD \cite{cheng2024spatialrgpt} & \textbf{99.2} & 99.0 & 80.2 & 92.0 & 87.5 & 91.8 & 91.6 & 41.2 & 65.6 & 51.9 & 49.6 & 57.9 & 53.2 & 72.4\\
         \circled{5} & SpatialRGPT-VILA-1.5-8B \cite{cheng2024spatialrgpt} & OSD \cite{cheng2024spatialrgpt} & \textbf{99.2} & \textbf{100.0} & 84.9 & 89.3 & 91.3 & 90.9 & 92.6 & 45.9 & 68.0 & 56.6 & 48.9 & 61.7 & 56.2 & 74.4\\
         \rowcolor{lightblue} 
         \circled{6} & MM1.5-3B \cite{zhang2024mm1} & -- & 35.8 & 46.7 & 44.3 & 52.7 & 47.1 & 50.9 & 46.3 & 4.7 & 10.7 & 2.8 & 1.5 & 12.0 & 6.4 & 26.3\\
         \rowcolor{lightgray} 
         \circled{7} & \Ours-3B & CA-VQA (our defs.) & 98.3 & 97.1 & 80.2 & 82.1 & 73.1 & 74.6 & 84.2 & 14.2 & 12.3 & 47.2 & 30.0 & 53.4 & 31.4 & 57.8\\
         \rowcolor{lightgreen} 
         \circled{8} & \Ours-3B & CA-VQA$^\star$ + OSD & 98.3 & 99.1 & \textbf{94.3} & 93.8 & 92.3 & 93.6 & 95.2 & 33.1 & 74.6 & 61.3 & 55.6 & 77.4 & 60.4 & 77.8\\
         \rowcolor{lightgreen} 
         \circled{9} & \Ours-3B + Depth (Tool; Mon.) & CA-VQA$^\star$ & 98.3 & 96.2 & \textbf{94.3} & 92.9 & 92.3 & \textbf{97.3} & 95.2 & 41.9 & 63.9 & 58.5 & 51.9 & 56.4 & 54.5 & 74.9\\
         \rowcolor{lightgreen} 
         \circled{10} & \Ours-3B + Depth (Tool; Mon.) & CA-VQA$^\star$ (scale aug.) & 98.3 & 98.1 & 92.5 & \textbf{95.5} & 92.3 & \textbf{97.3} & 95.7 & \textbf{60.8} & 72.1 & 58.5 & 48.1 & 58.7 & 59.6 & 77.7\\
         \rowcolor{lightgreen} 
         \circled{11} & \Ours-3B + Depth (Tool; Mon.) & OSD & 98.3 & \textbf{100.0} & 91.5 & 94.6 & \textbf{95.2} & 96.4 & 96.0 & 41.2 & 71.3 & 59.4 & 54.1 & 68.4 & 58.9 & 77.5\\
         \rowcolor{lightgreen} 
         \circled{12} & \Ours-3B + Depth (Tool; Mon.) & CA-VQA$^\star$ + OSD & 98.3 & 99.1 & 93.4 & 94.6 & 94.2 & \textbf{97.3} & \textbf{96.2} & 47.3 & \textbf{77.9} & \textbf{65.1} & \textbf{60.2} & \textbf{79.0} & \textbf{65.9} & \textbf{81.0}\\
         
         \midrule  
         \multicolumn{2}{c}{\textit{Specialist Models}}\\
         \midrule
         \rowcolor{lightgreen} 
         \rowcolor{lightgreen} 
         \circled{13} & \Ours-3B + Depth (Tool; Mon.) & CA-VQA$^\star$ + OSD & 98.3 & 98.1 & 93.4 & 94.6 & 93.3 & 96.4 & 95.7 & 59.5 & 82.0 & 63.2 & 55.6 & 83.5 & 68.7 & 82.2\\
         \rowcolor{lightgreen} 
         \circled{14} & \Ours-3B + Depth (Tool; Mon.) & CA-VQA$^\star$ & 98.3 & 94.3 & 93.4 & 90.2 & 89.4 & 95.5 & 93.5 & 37.8 & 61.5 & 63.2 & 51.9 & 56.4 & 54.2 & 73.8\\
         \circled{15} & & \emph{indoor} & & & & & & & & 50.0 & 90.2 & 63.2 & 60.0 & 71.8 & 67.0 &\\
         \circled{16} & & \emph{outdoor} & & & & & & & & 5.0 & 2.5 & -- & 0 & 3.3 & 2.7 & \\
         \rowcolor{lightgreen} 
         \circled{17} & \Ours-3B + Depth (Tool; Mon.) & CA-VQA$^\star$ (scale aug.) & 98.3 & 98.1 & 94.3 & 93.8 & 93.3 & 96.4 & 95.7 & 60.1 & 75.4 & 67.9 & 48.1 & 59.4 & 62.2 & 78.9\\
         \circled{18} & & \emph{indoor} & & & & & & & & 60.2 & 84.2 & 67.9 & 55.7 & 71.8 & 68.0 &\\
         \circled{19} & & \emph{outdoor} & & & & & & & & 60.0 & 57.5 & -- & 0 & 16.7 & 33.6 & \\
         \rowcolor{lightgreen} 
         \circled{20} & \Ours-3B + Depth (Tool; Mon.) & OSD & 98.3 & 100.0 & 89.6 & 92.9 & 92.3 & 96.4 & 94.9 & 39.2 & 73.0 & 59.4 & 53.4 & 82.0 & 61.4 & 78.1\\
         \circled{21} & & \emph{indoor} & & & & & & & & 32.4 & 76.8 & 59.4 & 54.8 & 78.6 & 60.4 & \\
         \circled{22} & & \emph{outdoor} & & & & & & & & 57.5 & 65.0 & -- & 44.4 & 93.3 & 65.1 & \\
         \rowcolor{lightgreen} 
         \circled{23} & \Ours-3B (Blind eval) & OSD & 100.0 & 98.1 & 74.5 & 81.3 & 86.5 & 80.1 & 86.9 & 21.6 & 22.1 & 43.4 & 27.1 & 21.8 & 27.2 & 57.0\\

         \bottomrule
    \end{tabular}
    }
    \caption{\textbf{SpatialRGPT-Bench Results.} \Ours-3B achieves SOTA with both image-only input and tool-use monocular depth, outperforming SpatialRGPT-VILA-1.5-8B (which fully encodes depth). Training on a mixture of CA-VQA$^\star$ and OSD performs best.}
    \label{tab:results_spatialrgpt}
\end{table*}

\noindent\textbf{\Ours's Metric Depth Estimation Ability.}
\cref{tab:results_depth_estimation} and \cref{fig:qualitative_example} show (quantitatively and qualitatively, respectively) that, surprisingly, our model's monocular metric depth estimation accuracy can even rival that of the SOTA DepthPro \cite{bochkovskii2024depth} specialist model.
While DepthPro is a general-purpose depth estimation model whereas \Ours is trained only on indoor scenes which aligns well with the CA-VQA benchmark\footnote{Note that DepthPro's training data mixture includes ARKitScenes \cite[App.\ C.1, Tab.\ 13]{bochkovskii2024depth} (which CA-VQA and thus \Ours's training data is based upon), so this is not a zero-shot evaluation for either model.}, these results still intriguingly suggest that MLLMs are capable of acquiring strong metric depth estimation abilities solely via data curation.

\subsection{CV-Bench Results}
\label{subsec:cv-bench_results}
The CV-Bench results in \cref{tab:results_cvbench} demonstrate that \Ours-3B \circled{10} significantly outperforms the much larger SOTA Cambrian-1-34B \circled{8}, highlighting the effectiveness of SFT on similar data. CoT \circled{11} and leveraging monocular (DepthPro) depth input via tool-use \circled{12} again further boost performance.
\Ours achieves almost perfect accuracy on the indoor splits of the 3D tasks, and also demonstrates strong out-of-domain generalization to the outdoor splits.
Notably, \Ours (Blind eval) \circled{13} achieves the best accuracy among all models on the 2D Object Count task, revealing a substantial bias in this benchmark. 
In contrast, on our CA-VQA benchmark, using vision input outperforms the blind baseline on \emph{Counting} by $\sim$13 points.

\subsection{SpatialRGPT-Bench Results}
\label{subsec:spatialrgpt-bench_results}
\cref{tab:results_spatialrgpt} shows SpatialRGPT-Bench results.
Notably, to align with the OpenSpatialDataset (OSD) used to train SpatialRGPT, SpatialRGPT-Bench is also based on \emph{axis-aligned} 3D boxes (AABBs), resulting in spatial concept definitions different to CA-VQA (see App.\ \ref{app:aabb_vs_obb}).
SpatialRGPT thus underperformed on CA-VQA, and similar issues arise when evaluating \Ours on their benchmark \circled{7}.
To enable a fair comparison of model capabilities, we thus align with the benchmark by generating \emph{CA-VQA$^\star$}, a variant of CA-VQA adopting their AABB-based definitions, and train \Ours on that.
We also train on OSD for comparison.
For \Ours + Depth we use monocular (DepthPro) metric depth via tool-use.
We observe the following:
\begin{itemize}[topsep=0pt, leftmargin=*]
    \item \textbf{Indoor vs.\ Outdoor.} While \Ours trained on CA-VQA$^\star$ \circled{9} achieves strong performance, it cannot significantly outperform the SOTA \circled{5}. Analysis reveals that while the model excels at indoor samples \circled{15}, it fails to generalize to outdoor samples \circled{16}, particularly on the \emph{Metric} tasks. We hypothesize that this is mainly attributed to the vast difference in metric scales between indoor and outdoor scenes, especially for object distances. We verify this with a simple \emph{scale augmentation} approach: we generate additional \emph{Distance} examples with scaling factors (sampled uniformly from $[1,10]$) applied to our underlying indoor scenes (i.e., 3D boxes and point clouds), resulting in a wider range of metric distances. We confirm that \Ours trained on CA-VQA$^\star$ + scale aug.\ \circled{17} substantially improves performance on the outdoor \emph{Distance} tasks \circled{19}, resulting in SOTA performance overall.
    \item \textbf{CA-VQA$^\star$ vs.\ OSD.} Training on OSD \circled{20} (based on diverse OpenImages \cite{kuznetsova2020open} data) results in strong performance both indoor \circled{21} and outdoor \circled{22}. However, training on CA-VQA$^\star$ still yields significantly better indoor performance overall \circled{15}, suggesting that our high-quality 3D GT is more effective than OSD's pseudo-annotations. When using CA-VQA$^\star$ + scale aug. \circled{19}, we become competitive with OSD \circled{22} even on outdoor \emph{Distance}, but still lack on outdoor \emph{Width / Height}.
    Combining CA-VQA$^\star$ with OSD \circled{13} results in significant improvements, emphasizing the complementary benefits of the two datasets.
    \item \textbf{\Ours vs.\ SpatialRGPT.} \Ours-3B outperforms the SOTA SpatialRGPT-VILA-1.5-8B \circled{5} with different data mixtures \circled{10}-\circled{12}, with and without \circled{8} depth input (SpatialRGPT uses depth maps via full encoding).
    \item \textbf{Depth vs.\ Image-only.} Leveraging monocular \emph{metric} depth via tool-use significantly improves performance for \Ours (\circled{12} vs.\ \circled{8}). SpatialRGPT-7B benefits less from fully encoding \emph{relative} depth (\circled{4} vs.\ \circled{3}).
    \item \textbf{Blind vs.\ Vision evaluation.} \Ours trained on OSD (Blind eval) \circled{23} performs well on several tasks, and GPT-4 \circled{1} was the prior SOTA for \emph{Width}. This suggests that SpatialRGPT-Bench and OSD suffer from significant biases and do not probe spatial perception alone.
\end{itemize}

\section{Conclusion}
\label{sec:conclusion}

We made several contributions towards unlocking object-centric 3D spatial understanding in MLLMs. First, we proposed a data generation pipeline, resulting in the CA-VQA SFT dataset for 3D perception tasks (incl.\ multi-view and depth inputs). Second, we introduced a new 3D spatial understanding benchmark, which includes tasks such as spatial relationships, metric estimation, and 3D grounding. Third, we demonstrated that our \Ours model can achieve SOTA performance on spatial reasoning benchmarks, while preserving general MLLM capabilities.
Lastly, we investigated how adding multi-view and depth as input modalities can further improve the model's spatial perception ability, and demonstrated that MLLMs can acquire strong monocular depth estimation capabilities via SFT.
In future work, we aim to extend our scope to outdoor scenes to complement our high-quality indoor dataset.

\newpage
\subsection*{Acknowledgments}
We would like to thank Anshul Shah, Lin Chen, Wei Liu, Ian Fasel, Alkesh Patel, Omer Hadad, Haoxuan You, Haotian Zhang, Wentao Wu, Philipp Dufter, Sergiu Sima, Sai Aitharaju, Albert Antony, David Haldimann, Michael Emmersberger, for helpful discussions and feedback.

{
    \small
    \bibliographystyle{ieeenat_fullname}
    \bibliography{main}
}
\appendix
\maketitlesupplementary
\setcounter{page}{1}

\begin{figure*}[t!]
\centering
\includegraphics[width=0.99\textwidth]{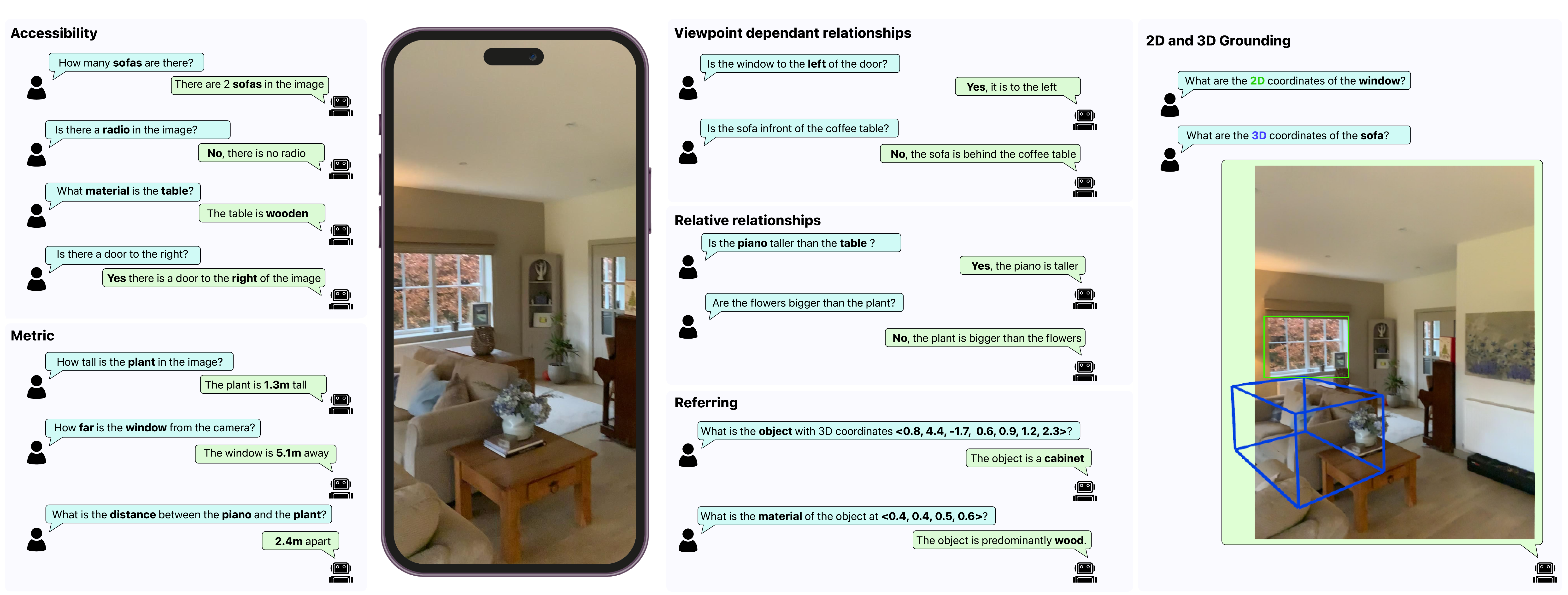}\vspace{-2mm}
\caption{
\textbf{CA-VQA Overview.} Example QA pairs from our Cubify Anything VQA (CA-VQA) dataset, aiming to unlock object-centric 3D spatial understanding in MLLMs. Using high-quality 3D ground truth annotations from CA-1M \cite{lazarow2024cubify}, we generate spatial perception questions across a variety of different tasks, e.g., involving \textbf{relative relationships}, \textbf{metric measurements}, and \textbf{3D object bounding boxes}.
}
\label{fig:qa_examples}
\end{figure*}

\section{More Details about the CA-VQA Data}
\label{app:data_details}

\subsection{Spatial Task Categories}

\begin{figure*}[t!]
\centering
\includegraphics[width=\textwidth]{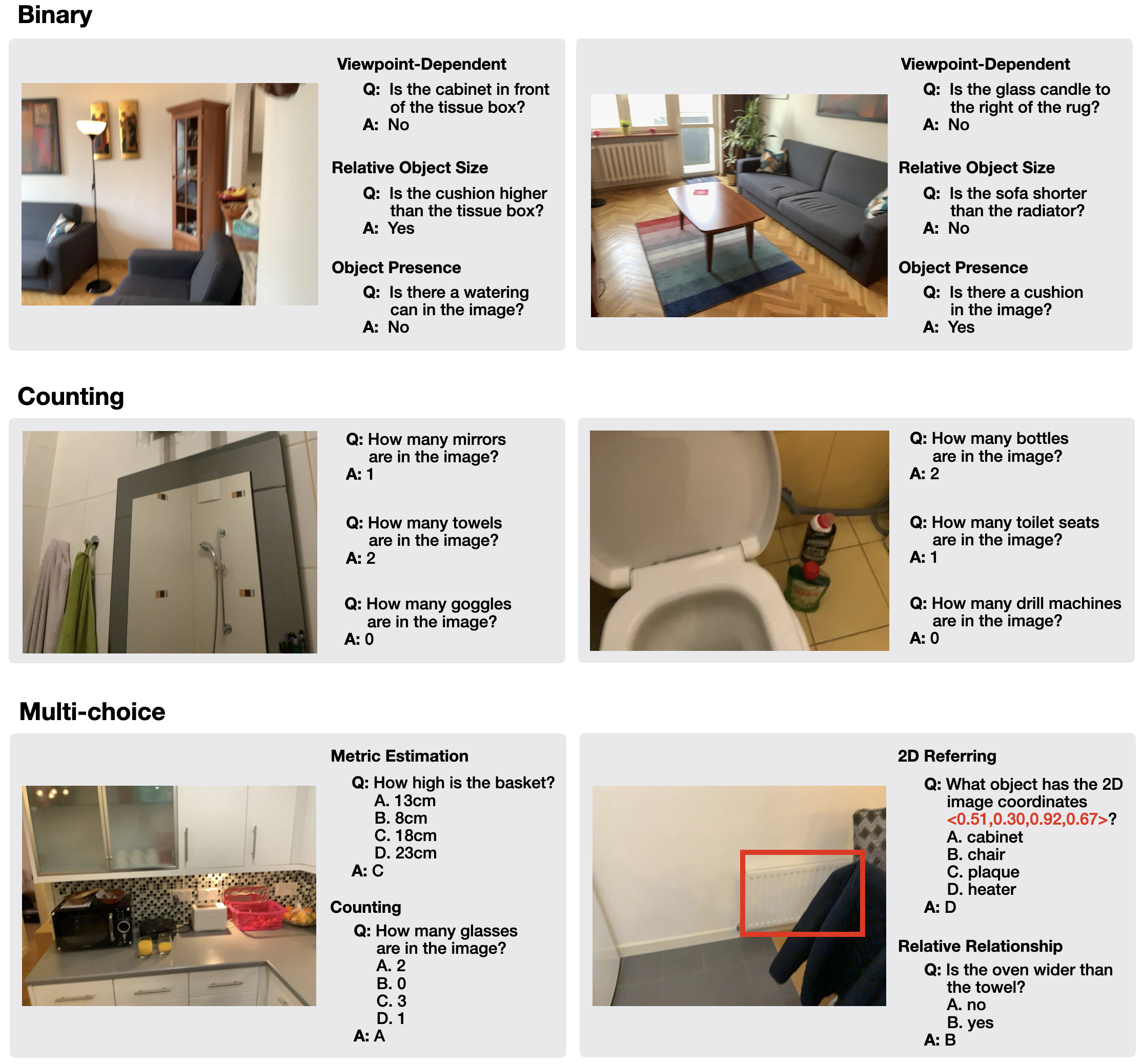}
\caption{Examples of CA-VQA data samples from the Binary, Counting and Multi-choice categories.}
\label{fig:spatial_category_examples_1}
\end{figure*}

\begin{figure*}[t!]
\centering
\includegraphics[width=\textwidth]{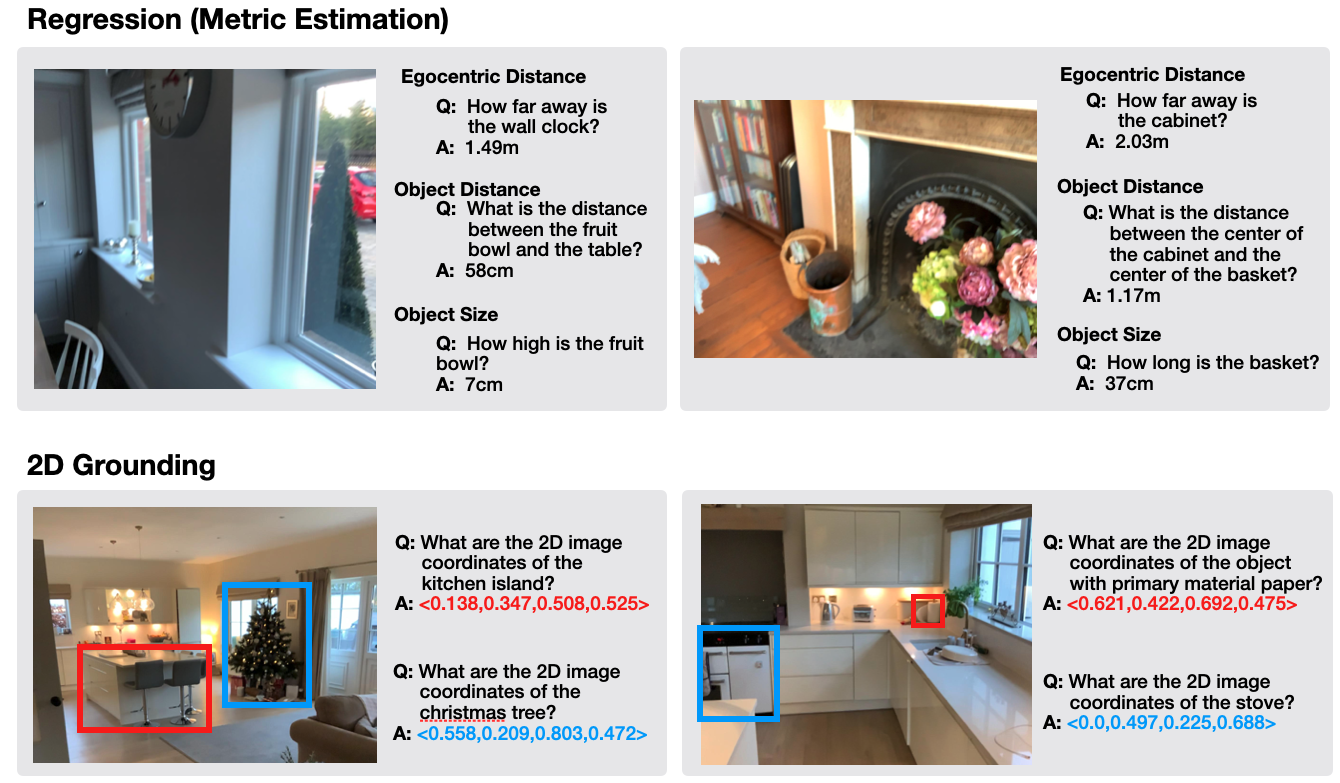}
\caption{Examples of CA-VQA data samples from the Regression (Metric Estimation) and 2D Grounding categories.}
\label{fig:spatial_category_examples_2}
\end{figure*}

We here provide more details on the different spatial task categories covered in CA-VQA, with visualizations of examples provided in \cref{fig:qa_examples,fig:spatial_category_examples_1,fig:spatial_category_examples_2}.
\begin{itemize}
    \item \textbf{Binary}.
    \begin{itemize}
        \item \textbf{Viewpoint-Dependent}. We consider the spatial relationships \emph{left vs.\ right} and \emph{in front vs.\ behind} between two objects, as determined from the current camera pose / viewpoint:\footnote{Note that we do not consider \emph{above vs.\ below} to avoid ambiguity: ``above'' could either refer to 2D image space (i.e., the 2D bounding box of A is above that of B), or to 3D space, where the latter can be ambiguous as well (i.e., do we just require that the 3D bounding box of A is located higher in terms of vertical dimension, or do we also require that A is located directly above B in terms of horizontal dimensions -- the latter might best match with how humans colloquially define ``above'').}
        \begin{itemize}
            \item \emph{Left vs.\ Right}. We determine the answer based on the horizontal coordinates of the objects' 2D bounding box centers.
            \item \emph{In front vs.\ Behind}. We determine the answer based on the distances between the camera and the objects' 3D bounding box centers.
        \end{itemize}
        \item \textbf{Relative Object Size}. We determine the answer based on the objects' \emph{width}, \emph{length} or \emph{height}, as defined in \textbf{Regression} below.
        \item \textbf{Object Presence}. For each sample asking about an object present in the image, we also generate a negative sample which asks about a (randomly sampled) object \emph{not} present in the image, to ensure a uniform distribution over answers (\emph{Yes / No}).
    \end{itemize}
    \item \textbf{Counting}. We determine the answer by simply counting the number of bounding boxes present in the image for a given object class. We also generate negative samples with (randomly sampled) objects not present in the image (i.e., such that the correct answer is 0).
    \item \textbf{Multi-choice}. This covers questions across the other spatial task categories, except for 2D and 3D grounding. We randomize the order of the options, obtaining the incorrect options as follows:
    \begin{itemize}
        \item \textbf{Regression (Metric Estimation)}. We compute three wrong options with either 10\% increments deviating from the real answer, or 5cm, whichever value is larger.
        \item \textbf{Counting}. We always ensure that 0 is an option (i.e., object is not present). We then randomly sample (additional) wrong options among the non-zero integers within $[\text{GT}-3, \text{GT}+3]$ (where GT is the correct answer), s.t.\ the total number of options is 4.
        \item \textbf{2D / 3D Referring}. We randomly sample three wrong object classes.
    \end{itemize}
    \item \textbf{Regression (Metric Estimation)}.
    \begin{itemize}
        \item \textbf{Egocentric Distance}. The distance between the camera and the \emph{closest} point of the object's point cloud.
        \item \textbf{Object Distance}. We consider both (minimum) distance and center distance between two objects:
        \begin{itemize}
            \item \emph{(Minimum) Distance}. The distance between the \emph{closest} points of the two objects' point clouds (i.e., minimum point distance).
            \item \emph{Center Distance}. The distance between the \emph{center} points of the two objects' 3D bounding boxes.
        \end{itemize}
        \item \textbf{Object Size}. We consider the 3D dimensions \emph{width}, \emph{length} and \emph{height}, defined as follows:
        \begin{itemize}
            \item \emph{Width.} The length of the \emph{larger} horizontal edge of the object's 3D bounding box (i.e., $\max(x_\text{len}, z_\text{len})$).
            \item \emph{Length.} The length of the \emph{shorter} horizontal edge of the object's 3D bounding box (i.e., $\min(x_\text{len}, z_\text{len})$)
            \item \emph{Height.} The length of the vertical edge of the object's 3D bounding box (i.e., $y_\text{len}$).
        \end{itemize}
    \end{itemize}
    \item \textbf{2D Grounding}. We use the 2D bounding box obtained from projecting the object's 3D bounding box into 2D image space.
    \item \textbf{3D Grounding}. We directly use the 3D bounding boxes provided in CA-1M \cite{lazarow2024cubify}.
\end{itemize}

\subsection{Depth: Chain-of-Thought (CoT) / Tool-Use}

We prepare multi-step CoT responses involving depth for questions within the \emph{Binary} (only ``behind vs.\ in front'') and \emph{Regression (Metric Estimation)} categories, as the ground truth answers for those rely on depth information. We also did preliminary experiments with \emph{3D Grounding} samples, but found that performance does not improve / even slightly regresses there, so we did not include any such samples in the final dataset.\footnote{We hypothesize that 3D grounding is too complex of a task to benefit from the simple depth information provided in the multi-step CoT answers, and that the model might just get confused. We leave a more comprehensive study of how to benefit 3D grounding with CoT for future work.}

The sequence format of the samples is illustrated in \cref{fig:depth_tool}, involving the target objects' 2D bounding boxes and depth values and the final (original) answer. We use the GT depth maps for generating the training examples, extracting the median depth value within the object's 2D bounding box\footnote{We also did preliminary experiments with other ways to extract a single depth value from the depth map within the 2D bounding box, such as the value at the center of the box or percentiles other than the median, but did not see significant improvements over using the median, which we found to be a robust choice.}. At test time, we then consider two alternative approaches for obtaining the depth values:
\begin{itemize}
    \item \textbf{Model prediction (CoT)}. We let the model predict the depth values (called \emph{CoT} in the experiments). As the model was trained on sequences involving the ground truth depth values, the models learn to predict depth. Our experiments reveal the accuracy of the resulting depth estimates.
    \item \textbf{Tool-use}. We allow the model to leverage a given depth map via tool-use. I.e., for a function call of the form $\text{Depth}(bbox) \rightarrow$ we extract the median depth value within the 2D bounding box, insert the depth value into the sequence, and then let the model continue its prediction to arrive at the final answer (see \cref{fig:depth_tool}).
\end{itemize}

\section{Optimal Data Mixture for \Ours}
\label{app:ablation_mixture_weights}

\begin{table*}[t!]
    \centering
    \resizebox{\linewidth}{!}{%
    \begin{tabular}{l|ll|cccc|cccc|c}
         \toprule
         \multirow{4}{*}{\textbf{Model}} & & & \multicolumn{9}{c}{Benchmark Category Averages}\\
         \cmidrule{2-12}
         & \multicolumn{2}{c|}{Mix.\ Ratio} & \multicolumn{4}{c|}{Spatial Understanding} & \multirow{3}{*}{General} & \multirow{3}{*}{Knowledge} & \multirow{3}{*}{Text-rich} & \multirow{3}{*}{Refer\&Ground} & \multirow{3}{*}{\textbf{Avg.}}\\
         \cmidrule{2-7}
         & Rel. & Eff. & CA-VQA & CV-Bench & SRGPT-Bench & \textbf{Avg.} & & & & &\\
         \midrule
         
         \rowcolor{lightblue} 
         MM1.5-3B \cite{zhang2024mm1} & 0:1 & 0:100 & 28.9 & 64.9 & 26.0 & 39.9 & 64.7 & 46.2 & 62.1 & 77.7 & 58.1\\
         \midrule
         & 1:1 & 12:88 & 66.3 & 91.2 & 52.8 & 70.1 & 65.0 & 46.2 & 62.1 & 79.1 & 64.5\\
         \rowcolor{lightgreen} 
         \Ours-3B & 2:1 & 22:78 & 67.1 & 92.4 & 53.7 & 71.1 & 64.8 & 46.7 & 61.4 & 78.8 & 64.5\\
         & 4:1 & 36:64 & 67.3 & 93.0 & 52.7 & 71.0 & 65.0 & 44.9 & 60.7 & 78.0 & 63.9\\
         & 8:1 & 54:46 & 67.4 & 93.1 & 53.7 & 71.4 & 64.8 & 46.8 & 61.2 & 79.0 & 64.6\\
         \midrule
         \rowcolor{lightgreen} 
         \Ours-3B & 1:0 & 100:0 & 67.1 & 93.0 & 54.1 & 71.4 & 42.6 & 34.7 & 17.2 & 23.9 & 38.0\\
         
         \bottomrule
    \end{tabular}
    }
    \caption{\textbf{Data Mixture Ratio Results.} Comparison of different data mixture ratios -- both (Rel)ative to the \emph{General} category (as in MM1.5), and (Eff)ective when considering the dataset sizes -- on aggregated metrics across the different benchmark categories.
    Overall, \Ours is a generalist MMLM that improves a lot on the \emph{Spatial} category while maintaining strong performance on the other categories.
    The data mixture ratio of 2:1 (spatial:general) provides a good performance trade-off and is used for \Ours throughout.
    The last line considers a spatial \emph{Specialist Model} that is trained on CA-VQA only; this model provides only a minor improvement on the spatial category, while regressing substantially on all other benchmark categories.
    }
    \label{tab:ablation_mixture_weights}
\end{table*}

We aim to build a generalist MLLM that excels across a variety of diverse tasks -- as opposed to a specialist that \emph{only} excels at spatial understanding. 
To this end, we identify the mixture weight for the new spatial data that achieves the best performance trade-off between the spatial vs.\ all other benchmark categories: general, knowledge, text-rich, 2D referring \& grounding.
Investigating the effect of adding a new model capability is particularly relevant for models with limited capacity, such as the 3B model we consider.

Results are shown in \cref{tab:ablation_mixture_weights}.
\Ours maintains similar performance as the MM1.5 baseline across most task categories, while significantly improving on the Spatial category.
This suggests that our model can successfully adopt the new spatial understanding capability without regressing on all the other capabilities, resulting in a generalist MLLM. 
The data mixture ratio of 2:1 (spatial:general) provides a good performance trade-off and is used for \Ours throughout.
We also consider a spatial \emph{Specialist Model} that is trained on CA-VQA only; however, this model provides only a small improvement on the spatial category, while regressing substantially on all other benchmark categories.
We use specialist models for some of our ablations to speed up experimentation.
\cref{app:additional_results} shows the detailed result breakdowns across the different task categories, compared to SOTA models.

\section{Results on Further Benchmark Categories}
\label{app:additional_results}

\begin{table*}[t!]
    \centering
    \resizebox{0.95\linewidth}{!}{%
    \begin{tabular}{l|ccc|cccccc}
         \toprule
         \multirow{4}{*}{\textbf{Model}}& \multicolumn{3}{c|}{Knowledge Benchmarks} & 
         \multicolumn{6}{c}{General Benchmarks}\\
         \cmidrule{2-10}
         &\makecell{AI2D\\(test)}&\makecell{MMMU\\(val)}&\makecell{MathV\\(testmini)}&\makecell{MME\\(P/C)}&SEED$^\text{I}$&POPE&LLaVA$^\text{W}$ & MM-Vet&RealWorldQA  \\
         
         \midrule
         MiniCPM-V 2.0-3B~\cite{yao2024minicpm}&62.9 & 38.2& 38.7 & 1808.2$^\dagger$&67.1 &87.8 & 69.2&38.2 &55.8 \\
         VILA1.5-3B~\cite{lin2024vila} &-- & 33.3 &-- &1442.4/-- &67.9&85.9 &--& --& --\\
         SpatialRGPT-VILA-1.5-3B~\cite{cheng2024spatialrgpt} &--  & 33.0 & -- & 1424.0/-- & 69.0 & 85.5 & -- & 38.2 & \\
         TinyLLaVA~\cite{zhou2024tinyllava}& --&  --& -- & 1464.9/--& --& 86.4& 75.8& 32.0 &-- \\
         Gemini Nano-2~\cite{team2023gemini} &51.0& 32.6 & 30.6& -- & -- & -- & --& --& --  \\
         Bunny~\cite{he2024efficient}  & -- & 41.4 &-- & 1581.5/361.1 & 72.5& 87.2&--&--&-- \\
         BLIP-3~\cite{xue2024xgen} & --& 41.1 & 39.6 & -- & 72.2 & 87.0 & -- & -- & 60.5 \\
         Phi-3-Vision-4B~\cite{abdin2024phi}& 76.7 & 40.4 & 44.5 & 1441.6/320.0& 71.8 & 85.8 & 71.6&	46.2&	59.4\\
         \rowcolor{lightblue} 
         MM1.5-3B \cite{zhang2024mm1} & 64.5 & 37.1 & 37.1 & 1423.7/277.9 & 70.2 & 87.9 & 74.3 & 37.1 & 57.7\\
         \rowcolor{lightgreen} 
         \Ours-3B & 63.6 & 36.6 & 38.4 & 1530.5/251.8 & 71.3 & 88.0 & 69.9 & 38.0 & 59.0\\
         
         \midrule
         \rowcolor{lavendermist} 
         Gemini-1.5-Pro~\cite{reid2024gemini} &79.1 & 60.6 & 57.7 & 2110.6$^\dagger$ & -- & 88.2 & 95.3 & 64.0 & 64.1 \\
         \rowcolor{lavendermist} 
         GPT-4V~\cite{openai2024gpt4vision} &75.9 & 53.8 & 48.7 & 1771.5$^\dagger$& 71.6 & 75.4 & 93.1 & 56.8 & 56.5\\
         \rowcolor{lavendermist} 
         GPT-4o~\cite{islam2024gpt}& 84.6 & 69.2 & 61.3 & 2310.3$^\dagger$& 77.1 & 85.6 & 102.0 & 69.1 & 75.4\\
         \bottomrule
    \end{tabular}
    }
    \caption{\textbf{Knowledge and General Benchmark Results.} Comparison with SOTA models on knowledge and general benchmarks. ($\dagger$) Sum of P and C scores. Gemini-1.5-Pro, GPT-4V and GPT-4o numbers are from \cite{duan2024vlmevalkit}.
    }
    \label{tab:results_general_knowledge}
\end{table*}

\begin{table*}[t!]
    \centering
    \resizebox{0.80\linewidth}{!}{%
    \begin{tabular}{l|ccccccc}
         \toprule
         \textbf{Model} &\makecell{WTQ\\(test)}&\makecell{TabFact\\(test)}&\makecell{OCRBench\\(test)}&\makecell{ChartQA\\(test)}&\makecell{TextVQA\\(val)}&\makecell{DocVQA\\(val)}&\makecell{InfoVQA\\(val)}\\
         \midrule

         MiniCPM-V 2.0-3B \cite{yao2024minicpm}& 24.2& 58.2& 60.5 & 59.8 & 74.1 & 71.9 & 37.6\\
         TinyLLaVA \cite{zhou2024tinyllava}&-- &-- &-- &-- & 59.1 & -- & -- \\
         Gemini Nano-2 \cite{team2023gemini} & --& --& -- & 51.9 & 65.9& 74.3 & 54.5\\
         BLIP-3-4B~\cite{xue2024xgen}& --& -- & --& -- & 71.0 & -- & -- \\
         Phi-3-Vision-4B \cite{abdin2024phi} &47.4&67.8&63.7& 81.4 & 70.1&83.3& 49.0  \\
         \rowcolor{lightblue} 
         MM1.5-3B \cite{zhang2024mm1} & 37.3 &	70.5 &	63.0 &	73.6 &	74.4 & 82.0 &	45.5\\
         \rowcolor{lightgreen}
         \Ours-3B &  36.2 & 71.0 & 60.0 &  75.0 & 75.3 & 82.7 & 43.7 \\

         \midrule
         \rowcolor{lavendermist} 
         Gemini-1.5-Pro~\cite{reid2024gemini}  &--& -- & 75.4 & 87.2 & 78.7 & 93.1 & 81.0 \\
         \rowcolor{lavendermist} 
         GPT-4V~\cite{openai2024gpt4vision} & -- & -- & 64.5 & 78.5$^\dagger$& -- & 88.4$^\dagger$ & --\\
         \rowcolor{lavendermist} 
         GPT-4o~\cite{islam2024gpt}& -- & -- & 73.6 & 85.7$^\dagger$ & -- & 92.8$^\dagger$ & --\\
         \bottomrule
    \end{tabular}
    }
    \caption{\textbf{Text-rich Benchmark Results.} Comparison with SOTA models on text-rich benchmarks.
    ($\dagger$) Numbers are obtained from \cite{li2024llava_onevision}.
    }
    \label{tab:results_textrich}
\end{table*}

\begin{table*}[t!]
    \centering
    \resizebox{0.70\linewidth}{!}{%
    \begin{tabular}{l|ccccc}
         \toprule
         \textbf{Model} &\makecell{RefCOCO\\(testA/B)}&	\makecell{RefCOCO+\\(testA/B)}&	\makecell{RefCOCOg\\(test)} &	\makecell{Flickr30k\\(test)}&	\makecell{LVIS-Ref\\(box/point)}\\

         \midrule
         MiniCPM-v2-3B \cite{yao2024minicpm}&  --  & -- & --  & -- & 48.2/47.7\\
         Phi-3-Vision-4B \cite{abdin2024phi}& 46.3 / 36.1 & 42.0 / 28.8 & 37.6   & 27.12 & 53.8/54.5\\
         InternVL2 \cite{chen2024far}& 88.2 / 75.9 & 82.8 / 63.3 & 78.3 & 51.6 & 51.0 / 51.1\\
         \rowcolor{lightblue} 
         MM1.5-3B \cite{zhang2024mm1} &91.7 / 85.7 & 87.67 / 75.23 & 85.9 & 85.1 & 74.0 / 58.2\\
         \rowcolor{lightgreen} 
         \Ours-3B & 92.2 / 85.9 & 88.3 / 76.8 & 86.8 & 85.1 & 75.9 / 58.5\\
         
         \bottomrule
    \end{tabular}
    }
    \caption{\textbf{2D Referring \& Grounding Benchmark Results.} Comparison with SOTA models on 2D referring and grounding benchmarks.}
    \label{tab:results_refer_ground}
\end{table*}

We here present a more detailed analysis of \Ours compared with SOTA baselines across the different benchmark categories.
Results on general and knowledge benchmarks are shown in \cref{tab:results_general_knowledge}, results on text-rich benchmarks are shown in \cref{tab:results_textrich}, and results on 2D referring \& grounding benchmarks are shown in \cref{tab:results_refer_ground}.
Overall, we observe that our \Ours model maintains a level of performance similar to the vanilla MM1.5 baseline.
This suggests that our model is able to successfully adopt the new spatial understanding capability without sacrificing performance on all the other model capabilities, resulting in a generalist MLLM.

\section{Analysis of Blind Filtering Procedure}
\label{app:blind_filtering_analysis}

\cref{tab:results_blind_filtering} analyses the effectiveness of our blind filtering procedure outlined in \cref{subsec:data_generation_pipeline} in ensuring that our CA-VQA benchmark becomes more reliant on vision input.
This is in contrast to some of the tasks from the other spatial understanding benchmarks we consider (CV-Bench and SpatialRGPT-Bench), where we found that blind models can perform very strongly and even rival models with vision input in some cases (see \cref{sec:experiments}).
Hence, these benchmarks would likely also benefit from blind filtering.

\begin{table*}[t!]
    \centering
    \resizebox{\linewidth}{!}{%
    \begin{tabular}{cll|rrrrrr|r}
         \toprule
         & \multirow{4}{*}{\textbf{Model}} & \multirow{4}{*}{\textbf{Eval Inputs}} & \multirow{3}{*}{Binary} & \multirow{3}{*}{Count.} & \multirow{3}{*}{Multi-c.} & \multicolumn{3}{c|}{Regression (Metric Estimation)} & \multirow{4}{*}{\textbf{Average}}\\
         \cmidrule{7-9}
         & & & & & & Ego-Dist. & Obj.-Dist. & Obj.-Size & \\
         \cmidrule{4-9}
         & & & Acc & Acc & Acc & \multicolumn{3}{c|}{Acc @ 10\% Relative Error ($\ell_1$)} & \\
         
         \midrule  
         \multicolumn{3}{c}{\textit{\textbf{Before} Blind Filtering}}\\
         \midrule
         \circled{1} & GPT-4 \cite{achiam2023gpt4} & Text & 57.9 & 35.1 & 52.7 & 8.9 & 8.2 & 17.0 & 30.0\\
         \circled{2} & GPT-4V~\cite{openai2024gpt4vision} & Image + Text & 61.6 & 68.1 & 63.2 & 6.4 & 8.4 & 19.7 & 37.9\\
         \rowcolor{lightgreen} 
         \circled{3} & Improvement from using vision & = \circled{2} -- \circled{1} & +3.7 & +33.0 & +10.5 & -2.5 & +0.2 & +2.7 & \textbf{+7.9}\\
         
         \midrule
         \circled{4} & \Ours-3B (Specialist) & Text & 69.3 & 69.5 & 77.6 & 12.9 & 11.0 & 25.2 & 44.3\\
         \circled{5} & \Ours-3B (Specialist) & Image + Text & 83.8 & 76.9 & 84.2 & 46.9 & 25.4 & 29.5 & 57.8\\
         \rowcolor{lightgreen} 
         \circled{6} & Improvement from using vision & = \circled{5} - \circled{4} & +14.5 & +7.4 & +6.6 & +34.0 & +14.4 & +4.3 & \textbf{+13.5}\\
         
         \midrule  
         \multicolumn{3}{c}{\textit{\textbf{After} Blind Filtering}}\\
         \midrule  
         \circled{7} & GPT-4 \cite{achiam2023gpt4} & Text & 9.6 & 8.5 & 9.6 & 6.2 & 6.2 & 5.8 & 7.7\\
         \circled{8} & GPT-4V~\cite{openai2024gpt4vision} & Image + Text & 39.2 & 63.3 & 32.9 & 11.4 & 9.3 & 10.1 & 27.7\\
         \rowcolor{lightgreen}
         \circled{9} & Improvement from using vision & = \circled{8} -- \circled{7} & +29.6 & +54.8 & +23.3 & +5.2 & +3.1 & +4.3 & \textbf{+20.0}\\
         
         \midrule  
         \circled{10} & \Ours-3B (Specialist) & Text & 34.3 & 60.8 & 60.7 & 10.1 & 8.4 & 17.9 & 32.0\\
         \circled{11} & \Ours-3B (Specialist) & Image + Text & 69.6 & 73.3 & 77.4 & 47.3 & 24.4 & 24.3 & 52.7\\
         \rowcolor{lightgreen}
         \circled{12} & Improvement from using vision & = \circled{11} -- \circled{10} & +35.3 & +12.5 & +16.7 & +37.2 & +16.0 & +6.4 & \textbf{+20.7}\\
         
         \midrule  
         \multicolumn{3}{c}{\textit{\textbf{Increase in Vision Improvement}: Before vs.\ After Blind Filtering}}\\
         \midrule  
         \rowcolor{lightgreen}
         \circled{13} & GPT-4/V & = \circled{9} -- \circled{3} & +25.9 & +21.8 & +12.8 & +7.7 & +2.9 & +1.6 & \textbf{+12.1}\\
         \rowcolor{lightgreen}
         \circled{14} & \Ours-3B (Specialist) & = \circled{12} -- \circled{6} & +20.8 & +5.1 & +10.1 & +3.2 & +1.6 & +2.1 & \textbf{+7.2}\\
         
         \bottomrule
    \end{tabular}
    }
    \caption{\textbf{CA-VQA Blind Filtering Analysis.} We study how the improvement from using vision (i.e., comparing a vision-evaluated model vs.\ a blind-evaluated model) changes after applying the blind filtering strategy outlined in \cref{subsec:data_generation_pipeline}, which follows \cite{chen2024mmstar}. Our results confirm that after applying our filtering strategy, 1) blind models perform substantially worse, and 2) vision improvements (i.e., the delta between vision and blind models) increase substantially, for both GPT-4/V and \Ours. This highlights the effectiveness of our blind filtering procedure in ensuring that our CA-VQA benchmark becomes more reliant on vision input (i.e., less susceptible to a strong language prior).}
    \label{tab:results_blind_filtering}
\end{table*}

\section{Axis-aligned vs.\ Oriented 3D Boxes}
\label{app:aabb_vs_obb}

\begin{figure*}[t!]
\centering
\includegraphics[width=\textwidth]{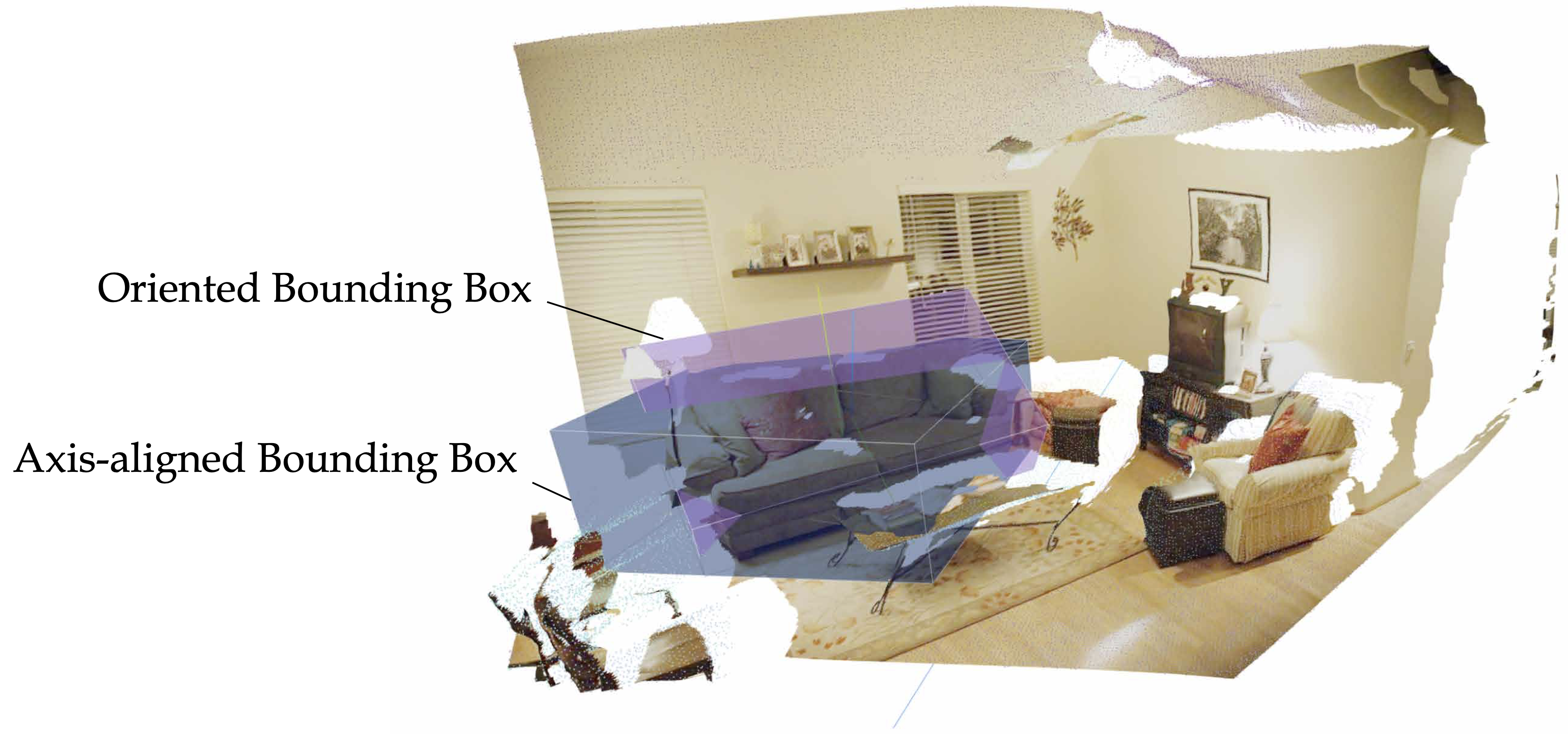}
\caption{Comparative visualization of axis-aligned vs.\ oriented 3D bounding boxes, taken from the SpatialRGPT paper \cite[Appendix K, Figure 11]{cheng2024spatialrgpt}. The object dimensions computed from AABBs can differ substantially from those computed from OBBs, depending on the object's rotation. For sake of illustration, assume that the sofa is 2m wide and 0.8m deep. We then obtain the following altered object dimensions when using an AABB instead of an OBB, at different yaw rotation angles (i.e., considering 7-DOF bounding boxes that are gravity-aligned / parallel to the ground, as in CA-1M / CA-VQA): width $\approx$ 2.1m and depth $\approx$ 1.7m with 30$^{\circ}$ rotation; width $\approx$ 1.7m and depth $\approx$ 2.1m with 60$^{\circ}$ rotation; and width = 0.8m and depth = 2m with 90$^{\circ}$ rotation (i.e., ``full'' rotation resulting in swapped dimensions).}
\label{fig:aabb_vs_obb}
\end{figure*}

\cref{fig:aabb_vs_obb} emphasizes the fundamental difference between axis-aligned (AABB) and oriented (OBB) 3D bounding boxes and how they affect the resulting object dimensions. This provides an indication of the misalignment issues arising when evaluating a model trained on data based on OBB ground truth (i.e., \Ours, which is based on the gravity-aligned 7-DOF yaw-oriented 3D bounding boxes from CA-1M) on a benchmark based on AABB ground truth (i.e., SpatialRGPT-Bench) and vice versa (i.e., evaluating SpatialRGPT on CA-VQA), as seen in \cref{subsec:ca-vqa_results,subsec:spatialrgpt-bench_results}.

\end{document}